\documentclass[10pt,twocolumn,letterpaper]{article}

\usepackage{cvpr}              % To produce the CAMERA-READY version
\usepackage{threeparttable}
\usepackage{booktabs}  
\usepackage{multirow}
\usepackage{algorithm}
\usepackage{algpseudocode}
\usepackage{etoolbox}
\usepackage{pifont}
\usepackage{wrapfig}
\usepackage{svg}

\definecolor{cvprblue}{rgb}{0.21,0.49,0.74}
\usepackage[pagebackref,breaklinks,colorlinks,allcolors=cvprblue]{hyperref}

\title{Effortless Active Labeling for Long-Term Test-Time Adaptation}

\author{Guowei Wang\\
South China University of Technology\\
{\tt\small eegw.wang@mail.scut.edu.cn}
\and
Changxing Ding\thanks{Corresponding author}\\
South China University of Technology\\
{\tt\small chxding@scut.edu.cn}
}

\begin{document}
\maketitle
\begin{abstract}
Long-term test-time adaptation (TTA) is a challenging task due to error accumulation. Recent approaches tackle this issue by actively labeling a small proportion of samples in each batch, yet the annotation burden quickly grows as the batch number increases. In this paper, we investigate how to achieve effortless active labeling so that a maximum of one sample is selected for annotation in each batch. First, we annotate the most valuable sample in each batch based on the single-step optimization perspective in the TTA context. In this scenario, the samples that border between the source- and target-domain data distributions are considered the most feasible for the model to learn in one iteration. Then, we introduce an efficient strategy to identify these samples using feature perturbation. Second, we discover that the gradient magnitudes produced by the annotated and unannotated samples have significant variations. Therefore, we propose balancing their impact on model optimization using two dynamic weights. Extensive experiments on the popular ImageNet-C, -R, -K, -A and PACS databases demonstrate that our approach consistently outperforms state-of-the-art methods with significantly  lower annotation costs. Code is available at: \href{https://github.com/flash1803/EATTA}{https://github.com/flash1803/EATTA}.
\end{abstract}    
\section{Introduction}
\label{sec:intro}

Test-time adaptation (TTA)~\cite{wang2020tent,liang2024comprehensive} aims to adapt pre-trained models to the new data distributions encountered during model deployment. TTA is crucial in dynamic scenarios such as autonomous driving and person re-identification~\cite{ding2020multi,tan2023style,tan2024harnessing,tan2024beyond,wang2022uncertainty}, where out-of-distribution data can significantly impair model performance. 
Recent approaches~\cite{wang2020tent,wang2022continual,chen2022contrastive, niu2022efficient} mostly resort to self-training techniques, such as pseudo-labeling and entropy minimization, to fine-tune source-trained models in the absence of ground-truth labels for the testing data. 
However, these approaches are prone to error accumulation and even result in negative transfer~\cite{wang2019characterizing}, especially during long-term distribution shifts~\cite{wang2022continual,gui2024active}.
\begin{figure}[tp]  
\vspace{-0.1in}
    \centering
    \includegraphics[width=1.0\linewidth]{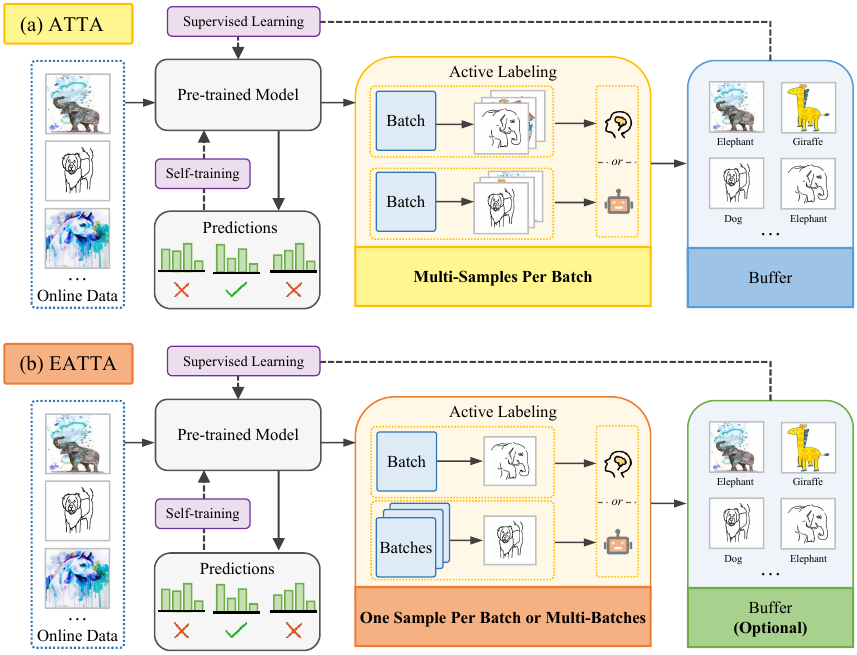}  
    \caption{Differences between (a) existing ATTA methods, and the proposed (b) EATTA. EATTA requires labeling only one sample per batch or multiple batches. And it does not rely on sample buffers. The solid and dashed black lines in the figure represent forward and backward propagation, respectively.}  
    \label{fig:eatta}  
\end{figure}

To facilitate long-term TTA, recent studies~\cite{gui2024active,chen2024towards,li2024exploring} propose the active test-time adaptation (ATTA), which introduces annotations created by experts~\cite{gui2024active,li2024exploring} or large foundation models~\cite{chen2024towards} during the model adaptation process. 
As illustrated in Figure~\ref{fig:eatta}, during each adaptation step, these methods label a small set of samples and then store them in a buffer, resulting in additional memory costs. They adapt the model according to the annotated samples in the buffer and the unlabeled online data. However, as the batch number increases, the data annotation cost rises significantly, affecting TTA efficiency.

The above observations motivate us to further relieve the annotation cost for ATTA. In particular, it is highly desirable to annotate only one sample per batch or even per multi-batches. However, this poses two major challenges: (i) how to identify the most valuable sample in each batch for labeling, and (ii) how to leverage this selected sample for model adaptation. First, existing ATTA methods, \textit{e.g.}, SimATTA~\cite{gui2024active}, tend to choose samples with high prediction entropy. However, samples with high prediction entropy are usually challenging for a model to learn via a single-step adaptation in the TTA context. Conversely, samples with low prediction entropy can hardly bring in effective adaptation. Second, existing works usually ignore the imbalance in gradient magnitudes between supervised and unsupervised objectives during adaptation~\cite{gui2024active,li2024exploring,chen2024towards}. Therefore, we aim to select samples that are informative and easy to learn, while ensuring balanced learning between supervised and unsupervised training objectives.

Herein, we propose an \textbf{E}ffortless \textbf{A}ctive labeling approach for long-term \textbf{TTA} (EATTA). First, we introduce a novel strategy to select the most valuable sample in each batch from the perspective of single-step optimization. Specifically, we regard the samples that borders between the source- and target-domain data distributions are considered the most feasible for the model to learn through single-step optimization. To achieve this, we obtain the pseudo-label for each sample in a batch according to its original feature. Then, we model the data distribution changes by adding slight noise to each sample's feature. After that, we recompute the prediction score on each sample's original pseudo-label, selecting the one with the largest prediction difference for annotation. We assume that if one sample borders between the two distributions, model predictions on its pseudo-label must be sensitive to slight data distribution changes. Finally, we perform supervised (\textit{i.e.}, cross-entropy loss) and unsupervised (\textit{i.e.}, entropy loss) learning on the actively labeled sample and the samples with confident pseudo-labels, respectively.

Second, we discover that the gradient magnitudes produced by these two training objectives are substantially different. This is because the prediction entropy for samples with confident pseudo-labels is already small. Therefore, the entropy loss results in small gradients. In comparison, we select a challenging sample for manual annotation and its pseudo-label is probably different from the manual annotation, allowing the cross-entropy loss to produce large gradients. This causes the model to overfit to the supervised objective. Hence, we propose a gradient norm-based debiasing method that balances the two objectives' impact with model optimization using two dynamic weights. Furthermore, we introduce an exponential moving average (EMA)-based strategy to refine the value of both weights, which further facilitates stable long-term TTA.

We demonstrate the effectiveness of our EATTA approach on popular test-time adaptation benchmarks, including ImageNet-C~\cite{hendrycks2019benchmarking}, -R~\cite{hendrycks2021many}, -K~\cite{wang2019learning}, -A~\cite{hendrycks2021nae}, and PACS~\cite{li2017deeper}. The results demonstrate that EATTA consistently outperforms current state-of-the-art ATTA methods with significantly fewer manual annotations. Notably, EATTA remains effective under limited annotation budgets, such as when the annotation number reduces to one sample per five batches on the ImageNet-C database.

\begin{figure*}[ht] 
\vspace{-0.1in}
    \centering
    \includegraphics[width=1.0\textwidth]{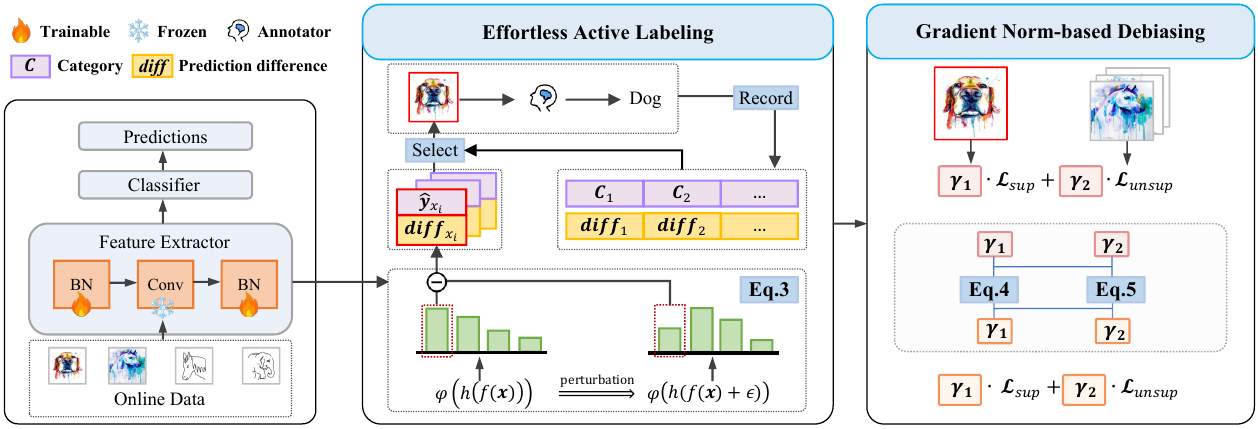} 
    \caption{\textbf{Overview of our EATTA approach}. 
    It aims to select at most one sample that is both informative and feasible to learn by a single-step optimization from each batch of data. We regard this sample lies at the border between the source- and target-domain data distributions, and identify this sample by observing its sensitivity to feature perturbations. Moreover, EATTA adopts a gradient norm-based debiasing strategy to adaptively combine the training objectives on the labeled and unlabeled data.}
    \label{fig:framework}
\end{figure*}

\section{Related Works}
\label{sec:related works}
\textbf{Test-Time Adaptation}.
Existing TTA methods tend to suffer from dramatic performance degradation during long-term distribution shifts~\cite{wang2022continual,brahma2023probabilistic,gui2024active}. This is mainly due to error accumulation caused by noisy pseudo-labels~\cite{chen2022contrastive}, catastrophic forgetting~\cite{wang2022continual,niu2022efficient}, and miscalibrated normalization statistics~\cite{hu2021mixnorm,su2024towards}.
To mitigate this, these approaches usually depend on reliable sample selection~\cite{niu2022efficient,lee2024entropy}, pseudo-label refinement~\cite{chen2022contrastive}, anti-forgetting techniques~\cite{wang2022continual,niu2022efficient,brahma2023probabilistic}, and robust prototypes~\cite{dobler2023robust,jang2022test,wang2025decoupled} during the self-training process. Additionally, calibration-based methods usually aim to estimate robust normalization statistics that incorporate target- and source-domain knowledge. To achieve this goal, they integrate source- and target-domain statistics through linear interpolation~\cite{mancini2018kitting,yuan2023robust,hu2021mixnorm,hong2023mecta,mirza2022norm}, update these statistics using a memory bank~\cite{gong2022note,yuan2023robust}, or adopt batch re-normalization~\cite{ioffe2017batch} to decouple gradient backpropagation~\cite{yang2022test,zhao2023delta}.

Recently, ATTA is proposed to handle long-term TTA by
introducing annotations from experts~\cite{gui2024active,li2024exploring} or large  models~\cite{chen2024towards}. 
For example, Gui \textit{et al}.~\cite{gui2024active} provided theoretical analysis, which demonstrate that introducing active learning improves TTA performance under distribution shifts. Li \textit{et al}.~\cite{li2024exploring} employed samples with human annotations for model selection.
Chen \textit{et al}.~\cite{chen2024towards} distilled knowledge from a large model to refine the edge-deployed model.
All these methods require human or large model annotators to label a small proportion of samples per batch. However, they disregard the growing cost of labeling that occurs as the batch number increases. In contrast, our EATTA approach provides a solution that requires annotating a maximum of one sample in each batch.

\noindent\textbf{Active Learning}. Active learning~\cite{li2024survey} adaptively selects informative samples for human annotation, reducing labeling and model training costs. Typically, it consists of uncertainty- and diversity-based methods. Uncertainty-based methods annotate samples where the model’s prediction exhibits a high uncertainty level, facilitating decision boundary refinement and improving performance in challenging cases. Popular metrics for assessing the uncertainty include least confidence~\cite{wang2014new}, maximum entropy~\cite{wang2014new}, minimum margin~\cite{xie2022active}, proximity to the decision boundary~\cite{tong2001support}, query-by-committee~\cite{seung1992query}, loss change~\cite{yoo2019learning}, and gradient norm magnitude~\cite{wang2022boosting}. 
Diversity-based methods aim to select samples that are representative of the broader data distribution, preventing the model from overfitting to limited data distributions. Therefor, they rely on clustering-~\cite{schohn2000less,nguyen2004active} and coreset-based 
sampling~\cite{sener2017active,kim2022defense}, and representative metrics~\cite{chakraborty2015active,li2022batch,ash2019deep} to achieve this goal. The main difference between common active learning and ATTA is that the former optimizes the model iteratively using a considerable amount of unlabeled data during the training phase. In contrast, ATTA focuses on online model adaptation during testing, utilizing limited computational resources for this process.

\section{Methods}

The proposed EATTA approach is illustrated in Figure~\ref{fig:framework}. In this section, We first introduce relevant notations and concepts in Section 3.1. We then address two key yet unsolved challenges for ATTA: (i) how to select the most valuable sample for labeling in Section 3.2, and (ii) how to leverage this selected sample for model adaptation in Section 3.3.

\subsection{Problem Statement}

TTA aims to adapt a model that has been pretrained on source domain data $\left\{\mathbf{x}_i, \mathbf{y}_i\right\}_{i=1}^{N_S} \in \mathcal{D}_S$ to the target domain $\left\{\mathbf{x}_j\right\}_{j=1}^{N_T} \in \mathcal{D}_T$ in an online manner without knowing ground-truth labels of the target domain data. $N_s$ and $N_t$ indicate the sample number in each domain, respectively. Parameters of the model are denoted as $\Theta$ and the model consists of a feature extractor $f(\cdot)$ and a classifier $h(\cdot)$.

For each online batch of data $\mathcal{B}^t$, the model first makes predictions for each of its samples 
through $\mathbf p(\mathcal{B}^t) = \phi(h(f(\mathcal{B}^t)))$, and then updates its parameters via the entropy loss:
\vspace{-0.1in}
 \begin{equation}\label{eqn.1}
\min _{\tilde{\Theta}} \mathbb{I}(\mathcal{B}^t) E(\mathcal{B}^t ; \Theta)=- \mathbb{I}(\mathcal{B}^t) \sum \mathbf p(\mathcal{B}^t) \log \mathbf p(\mathcal{B}^t),
\end{equation}
\vspace{-0.2in}

\noindent where $t$ stands for the batch index,  $\phi(\cdot)$ is the softmax operation, $E(\cdot)$ denotes entropy loss, and $\mathbb{I}(\cdot)$ is an indicator function. 
$\mathbb{I}(\cdot)=1$ indicates a sample with confident pseudo-label; otherwise, $\mathbb{I}(\cdot)=0$. Additionally, $\tilde{\Theta} \subseteq \Theta$ represents the subset of trainable parameters during TTA.

Compared with the common TTA setting, ATTA~\cite{gui2024active} allows the model to select a small subset of samples $\mathcal{B}_o^{\textit{t}}$ within $\mathcal{B}^\textit{t}$ for manual annotation. This enables the model to learn in a semi-supervised manner, and transforms the learning objective in Eq.~\ref{eqn.1} to:
\vspace{-0.05in}
\begin{equation}\label{eqn.2}
\mathcal{L}_{total} = \min _{\tilde{\Theta}} 
\gamma_1 \cdot CE(\mathcal{B}_o^{t} ; \Theta) +  \min _{\tilde{\Theta}}\gamma_2 \cdot \mathbb{I}(\mathcal{B}^t \setminus \mathcal{B}_o^{t})E( \mathcal{B}^t \setminus \mathcal{B}_o^{t} ; \Theta),
\end{equation}
% \vspace{-0.05in}
\noindent where $CE(\cdot)$ denotes the cross-entropy loss. $\gamma_1$ and $\gamma_2$ stand for the weights of the two loss terms, which are referred to as  $\mathcal{L}_{sup}$ and $\mathcal{L}_{unsup}$, respectively.

\subsection{Effortless Active Labeling}
Although existing ATTA methods annotate a small proportion of data in each batch, the actual annotation burden quickly grows as the batch number increases. This brings in difficulty for real-world application of ATTA. Therefore, it is highly desirable to annotate only the most valuable sample per batch or even per multi-batches. The common practice in active learning is to annotate samples with high prediction uncertainties, \textit{i.e.}, prediction entropy. However, this strategy is unsuitable to ATTA, as these samples are usually challenging for the model to learn by a single-step optimization in the TTA context.

\begin{figure}[tp] 
    \centering
    \includegraphics[width=1.0\linewidth, height=0.45\linewidth]{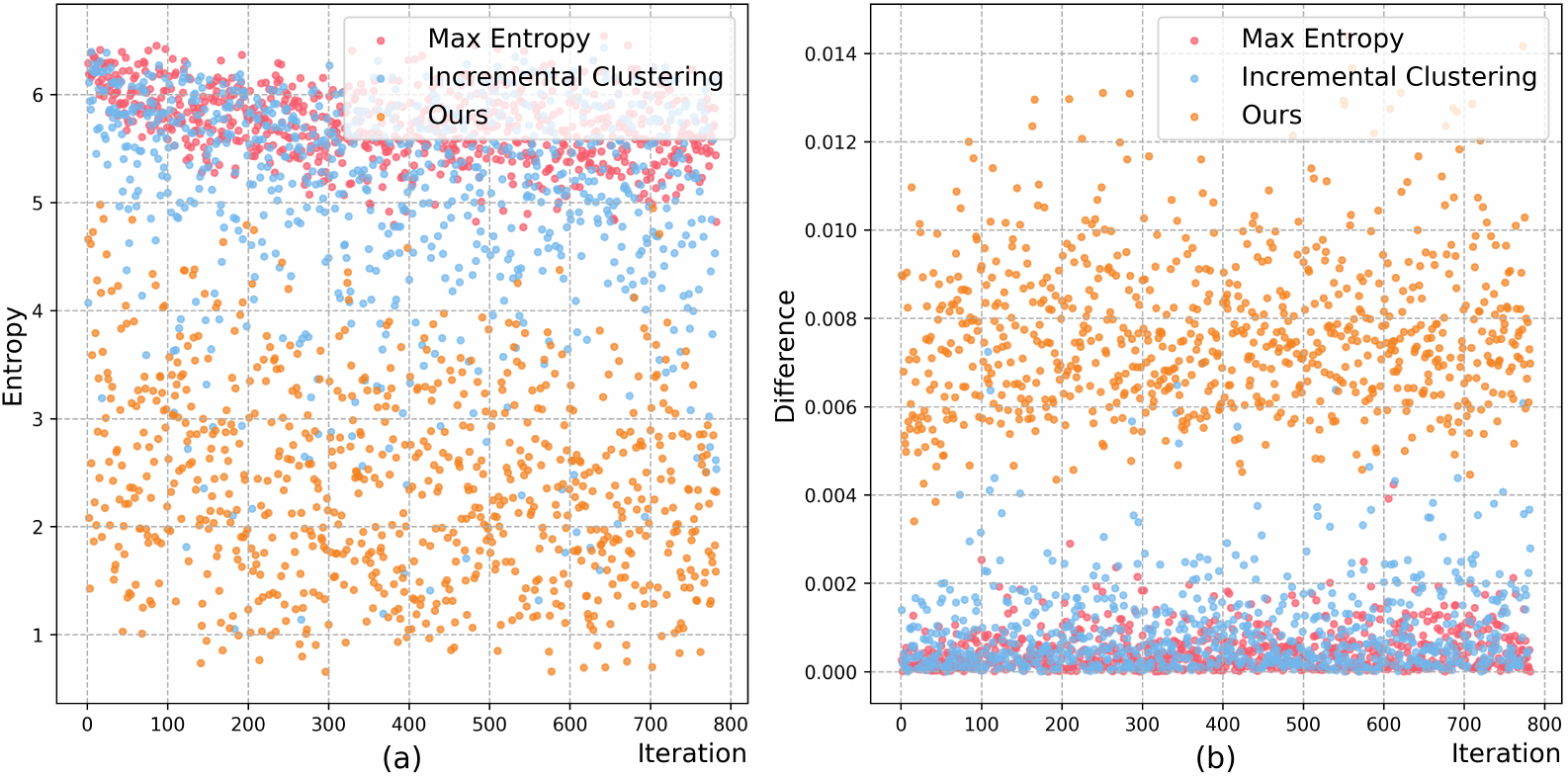} 
    \caption{(a) Distribution of the prediction entropy for the selected samples according to each of three criteria, respectively. (b) Distribution of the prediction confidence change after feature perturbation on the original pseudo-label of each selected sample. Three sample selection criteria are considered, \textit{i.e.}, maximum prediction entropy~\cite{wang2014new}, incremental clustering~\cite{gui2024active}, and ours. The experiments are conducted on the ImageNet-C database with severity level 5 gaussian noise.}
    \label{fig:ent_diff}
\end{figure}

Instead of labeling samples with high prediction entropy, existing ATTA methods, \textit{e.g.}, SimATTA~\cite{gui2024active}, adopt incremental clustering to select representative samples from each batch of target-domain data for manual annotation. However, as shown in Figure~\ref{fig:ent_diff}(a), the samples selected by incremental clustering still exhibit high prediction entropy, as they have to represent the target-domain data distribution that maybe quite different from that of the source domain. Conversely, selecting samples with low prediction entropy may provide limited information about the target domain, leading to ineffective adaptation. To solve this problem, we propose a criterion to identify samples that are both informative and easy to learn by assessing the robustness of the model to each sample.

Specifically, we regard the samples that lie at the border between the source- and target-domain data distributions as both informative and most feasible for the model to learn through single-step optimization. 
The evidence for this is that the border samples facilitate the model to gradually generalize to the target domain and therefore enhance its adaptation capacity 
~\cite{kumar2020understanding,chen2021gradual}. We also provide a toy example for demonstration in the supplementary material. 
Moreover, since they lie at the border between the two domains, the prediction confidence on their pseudo-labels must be sensitive to slight data perturbations. To efficiently model this data perturbation, we apply small Gaussian noise perturbations (\textit{i.e.}, $\epsilon \sim \mathcal{N}(\mu, \sigma^2)$) to their features, and measure the change in prediction confidence on the pseudo label of each sample:
\begin{equation}\label{Eqn.diff}
    \mathbf{diff}(\mathbf{x}_i) = \left|\phi (h(f(\mathbf{x}_i)))_{\hat{y}_{\mathbf{x}_i}} - \phi(h(f(\mathbf{x}_i)+ \epsilon))_{\hat{y}_{\mathbf{x}_i}}\right|,
    \vspace{-0.1in}
\end{equation}

\noindent where $\mathbf{x}_i$ denotes the $i$-th sample in the batch and $\hat{y}_{\mathbf{x}_i}$ denotes the pseudo label of this sample.

A large value of $\mathbf{diff}(\mathbf{x}_i)$ indicates that prediction for this sample is sensitive to slight changes in data distribution. As illustrated in Figure~\ref{fig:ent_diff}, the samples selected by our method are quite different from those chosen according to the maximum prediction entropy criterion~\cite{wang2014new} and the incremental clustering strategy~\cite{gui2024active}. In particular, the samples selected by our criterion exhibit noticeably lower prediction entropy, making them both informative and easier for the model to learn through single-step optimization.

Moreover, to avoid ineffective adaptation and lower the risk of model collapse due to the class imbalance problem~\cite{yuan2023robust}, we record the categories of the last $K$ annotated samples. We prioritize labeling one sample with the largest $\mathbf{diff}$ value among those whose pseudo labels are absent in the last $K$ annotations. If the pseudo-labels of all candidates overlap with recently recorded classes, then we simply choose the one with the largest $\mathbf{diff}$ value. 

\begin{figure}[tp] 
    \centering
    \includegraphics[width=1.0\linewidth, height=0.45\linewidth]{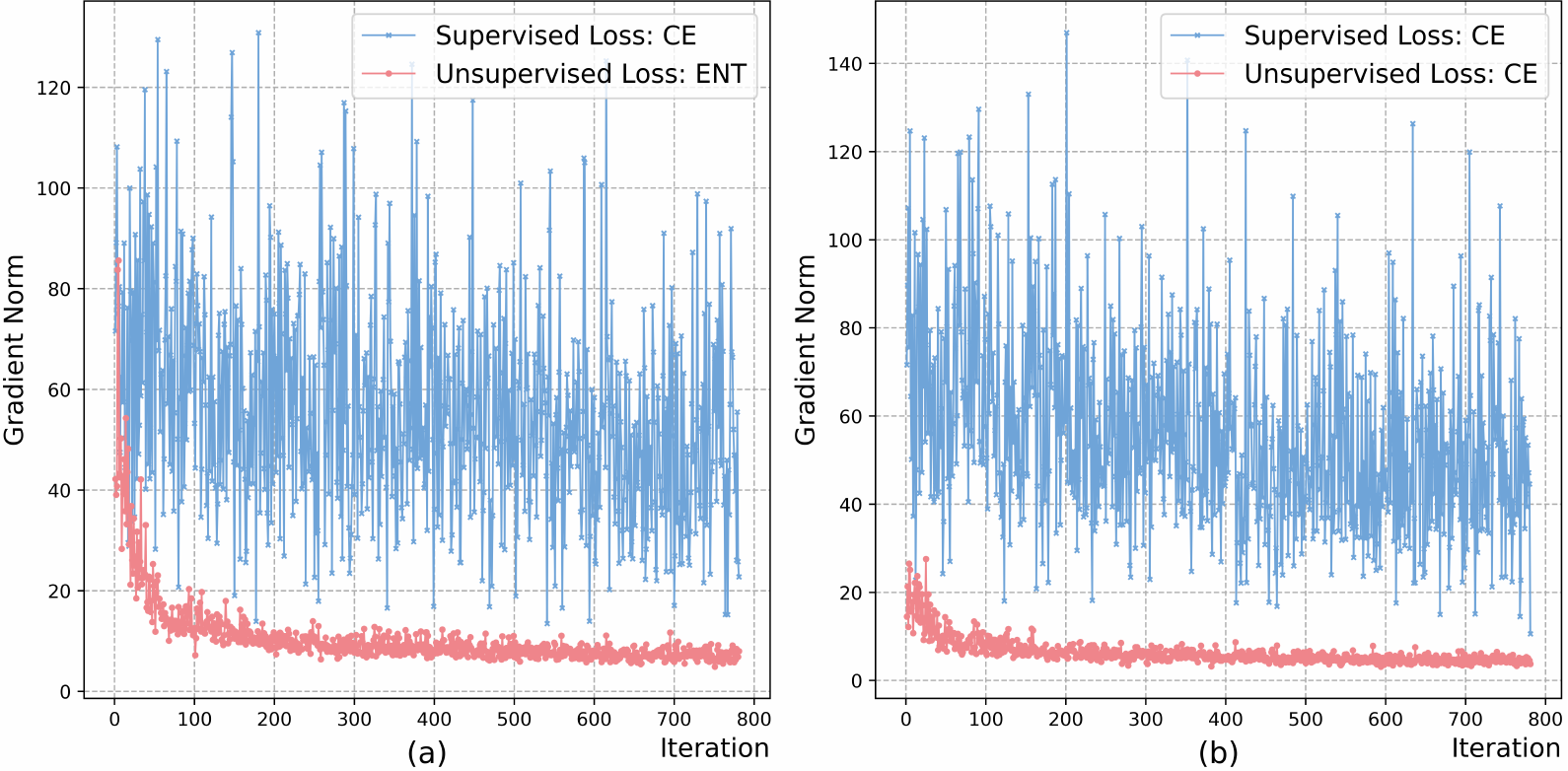} 
    \caption{
    (a) The gradient magnitudes of the supervised and unsupervised loss terms in Eq.~\ref{eqn.2}. (b) We repeat the former experiment while replacing the entropy loss with cross-entropy loss on the unannotated samples. 
    To employ this loss, we estimate pseudo-labels for the unannotated samples. 
    Both experiments are conducted on the ImageNet-C database with severity level 5 gaussian noise.
    }
    \label{fig:gradient-norm-bias}
\end{figure}

\subsection{Gradient Norm-based Debiasing}
Existing ATTA methods~\cite{gui2024active,li2024exploring,chen2024towards} often ignore the significant difference in gradient magnitude produced by the supervised and unsupervised training objectives. As illustrated in Figure~\ref{fig:gradient-norm-bias}, the gradient norm of $\mathcal{L}_{sup}$ is significantly larger than that of $\mathcal{L}_{unsup}$. This is mainly because $\mathcal{L}_{unsup}$ is imposed on samples with confident pseudo-labels, while $\mathcal{L}_{sup}$ is imposed on a sample whose pseudo-label is probably wrong. The imbalance between their gradient magnitudes brings in biased adaptation and causes the model to overfit to the single annotated sample.

To address this problem, we adopt a gradient norm-based debiasing strategy that balances the impact of these two training objectives via two dynamic weights.
Specifically, let the gradient norm for $\mathcal{L}_{sup}$ is $\|\nabla \mathcal{L}_{\text{sup}}\|_2=\sqrt{\sum_l\left(\frac{\partial \mathcal{L}_{\text{sup}}}{\partial {\tilde{\Theta}}_l}\right)^2}$ 
and that for $\mathcal{L}_{unsup}$ is $\|\nabla \mathcal{L}_{\text{unsup}}\|_2=\sqrt{\sum_l\left(\frac{\partial \mathcal{L}_{\text{unsup}}}{\partial{\tilde{\Theta}}_l}\right)^2}$, where $\tilde{\Theta}_l$ denotes the $l$-th trainable parameter. We then set the dynamic weights as:

\vspace{-0.1in}
\begin{equation}
\gamma_{1}=2 \cdot \left\|\nabla \mathcal{L}_{\text {unsup}}\right\|_2 / (\left\|\nabla \mathcal{L}_{\text {sup}}\right\|_2+\left\|\nabla \mathcal{L}_{\text {unsup}}\right\|_2),
\end{equation}
\begin{equation}
\gamma_{2}=2 \cdot \left\|\nabla \mathcal{L}_{\text {sup}}\right\|_2 / (\left\|\nabla \mathcal{L}_{\text {sup}}\right\|_2+\left\|\nabla \mathcal{L}_{\text {unsup}}\right\|_2).
\end{equation}

Furthermore, to stably adapt the model to long-term distribution shifts, we employ an exponential moving average (EMA)-based method to refine the value of $\gamma_1$ and $\gamma_2$:
\vspace{-0.1in}
\begin{equation}
\gamma_i^t \leftarrow \alpha \cdot \gamma_i^{(t-1)} + (1-\alpha) \cdot \gamma_i^{t}, i\in \{1,2\},
\end{equation}
% \vspace{-0.2in}
\noindent where $\alpha \in (0,1)$ is a trade-off parameter.

\begin{table*}[htp]
    \vspace{-0.1in}
    \caption{Performance comparisons on ImageNet-C in both CTTA and FTTA settings. The two settings are denoted as \textbf{C} and \textbf{F} in the tables, respectively. The backbone is ResNet-50 with BatchNorm. ``*'' and ``$\dagger$'' indicate 1 and 3 samples are annotated per batch, respectively. $\mathcal{BFS}$ represents the buffer size. 
    The \textbf{best} and \underline{second-best} performances are highlighted.}
    \vspace{-0.2in}
    \label{tab:imagenet-c}
    \newcommand{\tabincell}[2]{\begin{tabular}{@{}#1@{}}#2\end{tabular}}
    \begin{center}
    
    \begin{threeparttable}
        \resizebox{1.0\linewidth}{!}{
            \begin{tabular}{ll|ccc|cccc|cccc|cccc|c}
               & \multicolumn{1}{c}{}  & \multicolumn{3}{c}{Noise} & \multicolumn{4}{c}{Blur} & \multicolumn{4}{c}{Weather} & \multicolumn{4}{c}{Digital}  \\
                \toprule 
            \multicolumn{1}{l|}{\textbf{C}}  &  Methods  & Gauss. & Shot & Impul. & Defoc. & Glass & Motion & Zoom & Snow & Frost & Fog & Brit. & Contr. & Elastic & Pixel & JPEG & Avg. Err.  \\
               
                \midrule
        \multicolumn{1}{l|}{\multirow{5}{*}{\rotatebox{90}{Non-Active}}}      &  $\bullet$ Source  &97.8 &97.1 		 &98.2	 &82.1 	 &90.2 	 &85.2 	 &77.5 	 &83.1 	 &76.7 	 &75.6  &41.1 	 &94.6 		 &83.1 	 &79.0 	 &68.4  &82.0  \\

\multicolumn{1}{l|}{}& $\bullet~$TENT~\cite{wang2020tent}   &70.8 &63.9 		 &64.9	 &75.7 	 &75.1 	 &72.3 	 &65.3 	 &72.8 	 &75.7 	 &68.4  &56.0 	 &84.0 		 &73.1 	 &70.7 	 &74.9  &70.9
\\

\multicolumn{1}{l|}{}& $\bullet~$CoTTA~\cite{wang2022continual}   &78.2 &68.6 		 &64.3	 &75.2 	 &71.5 	 &69.6 	 &67.5 	 &72.0 	 &71.4 	 &67.2  &62.3 	 &73.5 		 &69.4 	 &66.8 	 &68.6  &69.8

\\
\multicolumn{1}{l|}{}&$\bullet~$SAR~\cite{niu2023towards}   &70.0 &62.2 		 &62.9	 &73.0 	 &70.1 	 &65.7 	 &58.0 	 &63.8 	 &64.1 	 &53.6  &42.0 	 &68.3 		 &53.8 	 &50.3 	 &53.2  &60.7

\\
\multicolumn{1}{l|}{}&$\bullet~$ETA~\cite{niu2022efficient}   &65.2 &59.5 		 &61.1	 &70.0 	 &69.0 	 &63.8 	 &57.3 	 &58.9 	 &60.8 	 &48.7  &39.5 	 &58.2 		 &48.6 	 &45.2 	 &48.1  &56.9
\\
\midrule

 \multicolumn{1}{l|}{\multirow{6}{*}{\rotatebox{90}{Active}}}&$\bullet~$Baseline$^{*}$  &69.3 &62.7 		 &64.4	 &72.9 	 &70.2 	 &64.7 	 &60.4 	 &63.8 	 &64.1 	 &53.3  &43.8 	 & 64.1		 &53.7	 &49.0 	 &52.3  &60.6
\\
\multicolumn{1}{l|}{}&$\bullet~$SimATTA$^\dagger$~\cite{gui2024active}  ($\mathcal{BFS}=300$)  &65.3 & 59.2		 &60.4	 &68.0 	 & 65.0	 & 58.4	 &53.0 	 & 54.9	 & 57.5	 &45.9  &38.0 	 & 56.8		 & 48.3	 &43.8 	 &47.2  &54.8
\\

\multicolumn{1}{l|}{}&$\bullet~$HILTTA$^\dagger$~\cite{li2024exploring}  &65.1 &57.6		 &58.5	 &65.5 	 &63.1 	 &56.9 	 &51.7 	 &54.4 	 &56.1	 &46.1  &36.9 	 &55.4 		 &47.2 	 &42.7 	 &46.2  &53.7
\\

\multicolumn{1}{l|}{} &\cellcolor{mycolor1}$\bullet~$Ours$^{*}$   &\cellcolor{mycolor1}64.9 &\cellcolor{mycolor1}57.5 	 &\cellcolor{mycolor1}57.9	 &\cellcolor{mycolor1}67.2 	 &\cellcolor{mycolor1}65.3 	 &\cellcolor{mycolor1}57.7 	 &\cellcolor{mycolor1}53.0 	 &\cellcolor{mycolor1}54.1 	 &\cellcolor{mycolor1}57.0 	 &\cellcolor{mycolor1}45.2  &\cellcolor{mycolor1}36.5 	 &\cellcolor{mycolor1}56.3 		 &\cellcolor{mycolor1}46.7 	 &\cellcolor{mycolor1}42.3 	 &\cellcolor{mycolor1}45.6  &\cellcolor{mycolor1}53.8
\\

\multicolumn{1}{l|}{}&\cellcolor{mycolor2}$\bullet~$\text{Ours}$^\dagger$   &\cellcolor{mycolor2}\underline{63.7} &\cellcolor{mycolor2}\underline{56.1} 		 &\cellcolor{mycolor2}\underline{56.6}	 &\cellcolor{mycolor2}\underline{65.2} 	 &\cellcolor{mycolor2}\underline{62.8} 	 &\cellcolor{mycolor2}\underline{54.4} 	 &\cellcolor{mycolor2}\underline{50.6} 	 &\cellcolor{mycolor2}\underline{52.2} 	 &\cellcolor{mycolor2}\underline{55.2} 	 &\cellcolor{mycolor2}\underline{43.6}  &\cellcolor{mycolor2}\underline{35.4} 	 &\cellcolor{mycolor2}\underline{52.9} 		 &\cellcolor{mycolor2}\underline{45.1} 	 &\cellcolor{mycolor2}\underline{40.8} 	 &\cellcolor{mycolor2}\underline{44.2}  &\cellcolor{mycolor2}\underline{51.9}
\\
\multicolumn{1}{l|}{}&\cellcolor{mycolor3}$\bullet~$\text{Ours}$^\dagger$ ($\mathcal{BFS}=300$)   &\cellcolor{mycolor3}\textbf{63.4} &\cellcolor{mycolor3}\textbf{55.7} 		 &\cellcolor{mycolor3}\textbf{55.7}	 &\cellcolor{mycolor3}\textbf{64.2} 	 &\cellcolor{mycolor3}\textbf{61.4}	 &\cellcolor{mycolor3}\textbf{53.1} 	 &\cellcolor{mycolor3}\textbf{49.3} 	 &\cellcolor{mycolor3}\textbf{50.4} 	 &\cellcolor{mycolor3}\textbf{53.9} 	 &\cellcolor{mycolor3}\textbf{42.3}  &\cellcolor{mycolor3}\textbf{34.6} 	 &\cellcolor{mycolor3} \textbf{52.2}		 &\cellcolor{mycolor3}\textbf{44.2} 	 &\cellcolor{mycolor3} \textbf{39.8}	 &\cellcolor{mycolor3}\textbf{43.4}  &\cellcolor{mycolor3}\textbf{50.9}

\\        
                \bottomrule
            \end{tabular}
        }
    \end{threeparttable}
    \end{center}

    \begin{center}
    \vspace{-0.1in}
    
    \begin{threeparttable}
        \resizebox{1.0\linewidth}{!}{
            \begin{tabular}{ll|ccc|cccc|cccc|cccc|c}
               & \multicolumn{1}{c}{}  & \multicolumn{3}{c}{Noise} & \multicolumn{4}{c}{Blur} & \multicolumn{4}{c}{Weather} & \multicolumn{4}{c}{Digital}  \\
                \toprule 
          \multicolumn{1}{l|}{\textbf{F}}    &  Methods  & Gauss. & Shot & Impul. & Defoc. & Glass & Motion & Zoom & Snow & Frost & Fog & Brit. & Contr. & Elastic & Pixel & JPEG & Avg. Err.  \\
               
                \midrule
              \multicolumn{1}{l|}{\multirow{5}{*}{\rotatebox{90}{Non-Active}}}   &$\bullet$ Source  &97.8 &97.1 		 &98.2	 &82.1 	 &90.2 	 &85.2 	 &77.5 	 &83.1 	 &76.7 	 &75.6  &41.1 	 &94.6 		 &83.1 	 &79.0 	 &68.4  &82.0  \\

\multicolumn{1}{l|}{}&$\bullet~$TENT~\cite{wang2020tent}   &70.8 &69.0 		 &70.0	 &71.9 	 &72.0 	 &58.3 	 &50.7 	 &52.6 	 &58.5 	 &42.4  &32.7 	 &70.3 		 &45.3 	 &41.5 	 &47.6  &56.9
\\

\multicolumn{1}{l|}{}&$\bullet~$CoTTA~\cite{wang2022continual}   &78.2 &77.8		 &77.2	 &81.8 	 &77.8 	 &63.8 	 &53.2 	 &57.6 	 &60.4 	 &44.1  &32.8 	 &73.5 		 &48.9 	 &43.0 	 &52.6  &61.5

\\
\multicolumn{1}{l|}{}&$\bullet~$SAR~\cite{niu2023towards}   &69.9 &69.2 		 &69.1	 &71.2 	 &71.7 	 &57.9 	 &50.8 	 &52.9 	 &57.8 	 &42.5  &32.7 	 &62.3 		 &45.7 	 &41.7 	 &47.8  &56.2

\\
\multicolumn{1}{l|}{}& $\bullet~$ETA~\cite{niu2022efficient}  &65.2 &62.4		 &63.5	 &66.8 	 &66.6 	 &52.6 	 &47.2 	 &48.4 	 &53.8 	 &40.2  &32.2	 & 54.8		 & 42.3	 &39.2 	 & 45.0 &52.0
\\
\midrule

\multicolumn{1}{l|}{\multirow{6}{*}{\rotatebox{90}{Active}}} &$\bullet~$Baseline$^{*}$  &69.3 &67.5 		 &68.1	 &72.6 	 &70.2 	 &57.7 	 &51.2 	 &52.3 	 &57.7 	 &43.0  &33.3 	 &63.0 		 &45.6 	 &42.1 	 &47.7  &56.1
\\
\multicolumn{1}{l|}{}&$\bullet~$SimATTA$^\dagger$~\cite{gui2024active} ($\mathcal{BFS}=300$)  &67.4 &63.7 		 &65.5	 &68.1 	 &66.8 	 &55.3 	 &49.7 	 &51.3	 &55.9 	 &42.7  &33.7 	 & 56.3		 &45.4	 &41.7 	 &47.4  &54.1
\\
\multicolumn{1}{l|}{}&$\bullet~$CEMA~\cite{chen2024towards} ($\mathcal{BFS}=300$)   &64.9 &69.1 		 &62.7	 &66.7 	 &66.5 	 &52.7 	 &48.4 	 &48.6 	 &54.6 	 &40.6  &33.5 	 &57.4 		 &43.2 	 &40.1 	 &45.1  &52.9
\\
\multicolumn{1}{l|}{}&$\bullet~$HILTTA$^\dagger$~\cite{li2024exploring}  &65.0 &63.1		 &64.5	 &66.1 	 &66.3 	 &54.4 	 &48.4 	 &49.8 	 &54.8	 &41.4  &32.5 	 &55.6 		 &43.6 	 &40.5 	 &45.8  &52.8
\\

\multicolumn{1}{l|}{}&\cellcolor{mycolor1}$\bullet~$Ours$^{*}$   &\cellcolor{mycolor1}64.9  &\cellcolor{mycolor1}62.1 	 &\cellcolor{mycolor1}63.4	 &\cellcolor{mycolor1}68.1	 &\cellcolor{mycolor1}66.9 	 &\cellcolor{mycolor1}51.8	 &\cellcolor{mycolor1}47.4 	 &\cellcolor{mycolor1}48.1 	 &\cellcolor{mycolor1}54.3 	 &\cellcolor{mycolor1}40.4  &\cellcolor{mycolor1}31.7 	 &\cellcolor{mycolor1}61.4 		 &\cellcolor{mycolor1}\underline{42.7} 	 &\cellcolor{mycolor1}39.2 	 &\cellcolor{mycolor1}44.8  &\cellcolor{mycolor1}52.4
\\

\multicolumn{1}{l|}{}&\cellcolor{mycolor2}$\bullet~$\text{Ours}$^\dagger$    &\cellcolor{mycolor2}\underline{63.7} &\cellcolor{mycolor2}\underline{61.4}		 &\cellcolor{mycolor2}\underline{63.1}	 &\cellcolor{mycolor2}\underline{65.8} 	 &\cellcolor{mycolor2}\underline{65.7} 	 &\cellcolor{mycolor2}\underline{50.6} 	 &\cellcolor{mycolor2}\underline{47.0} 	 &\cellcolor{mycolor2}\underline{47.5} 	 &\cellcolor{mycolor2}\underline{53.7} 	 &\cellcolor{mycolor2}\underline{40.0}  &\cellcolor{mycolor2}\underline{31.6} 	 &\cellcolor{mycolor2}\underline{55.9}		 &\cellcolor{mycolor2}43.0 	 &\cellcolor{mycolor2}\underline{38.8} 	 &\cellcolor{mycolor2}\underline{44.6}  &\cellcolor{mycolor2}\underline{51.5}
\\
\multicolumn{1}{l|}{}&\cellcolor{mycolor3}$\bullet~$\text{Ours}$^\dagger$ ($\mathcal{BFS}=300$)   &\cellcolor{mycolor3}\textbf{63.3}  &\cellcolor{mycolor3}\textbf{60.6} 		 &\cellcolor{mycolor3}\textbf{61.9}	 &\cellcolor{mycolor3}\textbf{64.7}	 &\cellcolor{mycolor3}\textbf{64.5}	 &\cellcolor{mycolor3}\textbf{50.0} 	 &\cellcolor{mycolor3}\textbf{46.6} 	 &\cellcolor{mycolor3}\textbf{47.2} 	 &\cellcolor{mycolor3}\textbf{52.9} 	 &\cellcolor{mycolor3}\textbf{39.6}  &\cellcolor{mycolor3}\textbf{31.3} 	 &\cellcolor{mycolor3}\textbf{53.8}		 &\cellcolor{mycolor3}\textbf{42.0} 	 &\cellcolor{mycolor3}\textbf{38.6} 	 &\cellcolor{mycolor3}\textbf{44.0}  &\cellcolor{mycolor3}\textbf{50.7}

\\        
                \bottomrule
            \end{tabular}
        }
    \end{threeparttable}
    \end{center}
\end{table*}

\section{Experiments}

\noindent\textbf{Baselines}. We compare the performance of EATTA with four state-of-the-art TTA methods, including TENT~\cite{wang2020tent}, COTTA~\cite{wang2022continual}, ETA~\cite{niu2022efficient}, and SAR~\cite{niu2023towards}; as well as three ATTA methods, namely SimATTA~\cite{gui2024active}, CEMA~\cite{chen2024towards}, and HILTTA~\cite{li2024exploring}. Note that all methods have no access to source-domain data to facilitate a fair comparison. Additionally, we build an ATTA baseline that randomly selects a specified number of samples from each online batch for manual annotation, and perform ATTA using Eq.~\ref{eqn.2}. Previous ATTA methods~\cite{gui2024active,chen2024towards} usually store the manually annotated samples in a buffer and sample 32 images per adaptation step to perform supervised learning. We fix the buffer size $\mathcal{BFS}$ to 300 for fair comparison.

\noindent\textbf{Datasets}.
We evaluate the performance of all methods using the dataset with synthetic corruptions, \textit{e.g.}, ImageNet-C~\cite{hendrycks2019benchmarking}, which comprises 15 types of corruptions at 5 severity levels. In line with previous work~\cite{wang2020tent,wang2022continual,niu2022efficient,niu2023towards,li2024exploring}, we use level
5 severity for testing. We also conduct evaluations on datasets with real-world distribution shifts, such as ImageNet-R~\cite{hendrycks2021many}, ImageNet-K~\cite{wang2019learning}, ImageNet-A~\cite{hendrycks2021nae}, and PACS~\cite{li2017deeper}. Details of these datasets are provided in the supplementary material.

\noindent\textbf{Backbones}. We conduct experiments on ResNet-50 with BatchNorm~\cite{santurkar2018does} and GroupNorm~\cite{wu2018group}, and ViT-B-16 with LayerNorm for ImageNet-C/R/K/A. 
The pretrained models are provided by Pytorch and timm library.
For experiments on PACS, we adopt ResNet-18 with BatchNorm as the backbone and pretrain the source models following the official training protocol~\cite{gui2024active, gulrajani2020search}.

\noindent\textbf{Settings}.
We adopt evaluation settings for both continual test-time adaptation (CTTA)~\cite{wang2022continual} and fully test-time adaptation (FTTA)~\cite{wang2020tent}. Their key difference is that whether the model can restore its parameters when distribution shift changes. 
Existing ATTA methods, \textit{e.g.}, SimATTA~\cite{gui2024active} and CEMA~\cite{chen2024towards}, focus on the FTTA setting, while HILTTA~\cite{li2024exploring} addresses the CTTA setting. 
In contrast,
we demonstrate the robustness of our method in both settings. In the following tables, unless otherwise specified, the symbols `*' and `$\dagger$' indicate that 1 and 3 samples are annotated per batch, respectively. And the `Source' denotes the performance of the pretrained source model without any adaptation. Moreover, we denote CTTA and FTTA as `\textbf{C}' and `\textbf{F}' for simplicity, respectively.

\noindent\textbf{Implementation Details}.
We set the batch size to 64 for all experiments. Following~\cite{niu2022efficient,niu2023towards}, the indicator function in Eq.~\ref{eqn.1} outputs 1 if the prediction entropy for a sample is below $0.4 \times \ln(\mathcal{C})$, where $\mathcal{C}$ is the number of classes. The 
\begin{table*}[htp]
\vspace{-0.1in}
    \caption{Performance comparisons on ImageNet under the FTTA setting. `RN50-BN' and `RN50-GN' denote the ResNet50 model with BatchNorm and GroupNorm, respectively. 
    }
    \vspace{-0.15in}
    \label{tab:imagenet-variants}
\resizebox{1.0\linewidth}{!}{
\begin{tabular}{ll|cccc|cccc|cccc}
\multicolumn{1}{c}{} 
 &\multicolumn{1}{l}{}  & \multicolumn{4}{c}{ImageNet-R}  & \multicolumn{4}{c}{ImageNet-K} & \multicolumn{4}{c}{ImageNet-A}                                                                   \\ \toprule
\multicolumn{1}{l|}{\textbf{F}}&\multicolumn{1}{l|}{Methods}& \multicolumn{1}{c}{RN50-BN} & \multicolumn{1}{c}{RN50-GN} & \multicolumn{1}{c}{ViT-B-16} & \multicolumn{1}{c|}{Avg. Err.}  & \multicolumn{1}{c}{RN50-BN} & \multicolumn{1}{c}{RN50-GN} & \multicolumn{1}{c}{ViT-B-16}  & \multicolumn{1}{c|}{Avg. Err.} & \multicolumn{1}{c}{RN50-BN} & \multicolumn{1}{c}{RN50-GN} & \multicolumn{1}{c}{ViT-B-16}  & \multicolumn{1}{c}{Avg. Err.} \\ \midrule

\multicolumn{1}{l|}{\multirow{5}{*}{\rotatebox{90}{Non-Active}}}&$\bullet~$Source    &63.8    &59.3    &56.0  &59.7  &75.9   &70.5    &70.6  &72.3  &100.0  &86.7  &79.2   &88.6 \\
\multicolumn{1}{l|}{}&$\bullet~$TENT~\cite{wang2020tent}   &57.8   &56.9    &53.4  &56.0  &69.5  &68.7    &65.6  &66.9 &99.9   &86.2   &77.4  &87.8 \\
\multicolumn{1}{l|}{}&$\bullet~$CoTTA~\cite{wang2022continual}   &57.3   &58.6    &55.4  &57.1  &69.9   &73.0   &98.2  &80.4  &99.8  &89.2   &79.3   &89.4 \\
\multicolumn{1}{l|}{}&$\bullet~$SAR~\cite{niu2023towards}   &57.2   &57.4    &48.8   &54.5  &68.5   &68.7  &70.4   &69.2  &99.9   &86.2  &74.9   &87.0 \\
\multicolumn{1}{l|}{}&$\bullet~$ETA~\cite{niu2022efficient}    &54.0   &55.8  &48.8  &52.9  &64.3  &66.0  &59.4   &63.2 &99.8  &86.2  &75.9  &87.3\\
\midrule
\multicolumn{1}{l|}{\multirow{7}{*}{\rotatebox{90}{Active}}}&$\bullet~$Baseline$^{*}$  &54.9   &54.6  &48.5  &52.7  &66.9  &62.4  &59.9  &63.1  &99.4   &85.2   &76.5  &87.0  \\ 
\multicolumn{1}{l|}{}&$\bullet~$SimATTA$^\dagger$~\cite{gui2024active} ($\mathcal{BFS}=300$)  &51.3  &50.3  & 45.1  &48.9  &64.0  &58.6  &57.2  &59.9  &97.2   &85.0   &72.4  &84.9  \\ 
\multicolumn{1}{l|}{}&$\bullet~$CEMA~\cite{chen2024towards} ($\mathcal{BFS}=300$)  &51.4  &49.6  &44.6  &48.5  &65.6  &61.2  &60.0  &62.3  &97.7   &82.0   &72.9  &84.2  \\ 
\multicolumn{1}{l|}{}&$\bullet~$HILTTA$^\dagger$~\cite{li2024exploring} &52.6  &46.9  &43.9  &47.8  &63.3  &57.2  &58.1  &59.5  &98.3   &80.0   &72.2  &83.5  \\
\multicolumn{1}{l|}{}&\cellcolor{mycolor1}$\bullet~$Ours$^{*}$  &\cellcolor{mycolor1}51.5  &\cellcolor{mycolor1}46.7  &\cellcolor{mycolor1}44.5  &\cellcolor{mycolor1}47.6  &\cellcolor{mycolor1}64.3  &\cellcolor{mycolor1}57.3 &\cellcolor{mycolor1}58.3  &\cellcolor{mycolor1}60.0  &\cellcolor{mycolor1}98.1 &\cellcolor{mycolor1}83.4  &\cellcolor{mycolor1}72.1  &\cellcolor{mycolor1}84.5 \\ 
\multicolumn{1}{l|}{}&\cellcolor{mycolor2}$\bullet~$Ours$^\dagger$  &\cellcolor{mycolor2}\underline{49.7}   &\cellcolor{mycolor2}\underline{45.9}  &\cellcolor{mycolor2}\underline{43.8}  &\cellcolor{mycolor2}\underline{46.5}  &\cellcolor{mycolor2}\underline{62.4}   &\cellcolor{mycolor2}\underline{56.2}  &\cellcolor{mycolor2}\underline{57.6}  &\cellcolor{mycolor2}\underline{58.7}  &\cellcolor{mycolor2}\underline{97.2}  &\cellcolor{mycolor2}\underline{79.3}  &\cellcolor{mycolor2}\underline{71.7}  &\cellcolor{mycolor2}\underline{82.8} \\
\multicolumn{1}{l|}{}&\cellcolor{mycolor3}$\bullet~$Ours$^\dagger$ ($\mathcal{BFS}=300$)  &\cellcolor{mycolor3}\textbf{48.8}  &\cellcolor{mycolor3}\textbf{45.3}  &\cellcolor{mycolor3}\textbf{43.1} &\cellcolor{mycolor3}\textbf{45.7} &\cellcolor{mycolor3}\textbf{61.4} &\cellcolor{mycolor3}\textbf{55.8}   &\cellcolor{mycolor3}\textbf{56.5}  &\cellcolor{mycolor3}\textbf{57.9}  &\cellcolor{mycolor3}\textbf{96.1}  &\cellcolor{mycolor3}\textbf{78.4}    &\cellcolor{mycolor3}\textbf{70.9}   &\cellcolor{mycolor3}\textbf{81.8}  \\
\bottomrule
\end{tabular} 
}
\end{table*}
\begin{table*}[t]
    \setlength{\tabcolsep}{4pt}
    \caption{Performance comparisons on PACS under the CTTA and FTTA settings. The two settings are denoted as C and F in the table, respectively.
    }
    \vspace{-0.25in}
    \label{tab:pacs}
\begin{center}
 \resizebox{1.0\linewidth}{!}{
 
\begin{tabular}{ll|cccccccc|cccccccc|cccccccc|cccccccc}
\toprule

\multicolumn{1}{l|}{\textbf{C}}&\multicolumn{1}{l|}{Methods} &
\multicolumn{1}{c}{P} & \multicolumn{1}{@{}c@{}}{$\rightarrow$} & \multicolumn{1}{c}{A} & \multicolumn{1}{@{}c@{}}{$\rightarrow$} & \multicolumn{1}{c}{C} & \multicolumn{1}{@{}c@{}}{$\rightarrow$} & \multicolumn{1}{c}{S} & \multicolumn{1}{c}{Avg. Acc.} &

\multicolumn{1}{|c}{A} & \multicolumn{1}{@{}c@{}}{$\rightarrow$} & \multicolumn{1}{c}{C} & \multicolumn{1}{@{}c@{}}{$\rightarrow$} & \multicolumn{1}{c}{P} & \multicolumn{1}{@{}c@{}}{$\rightarrow$} & \multicolumn{1}{c}{S} & \multicolumn{1}{c|}{Avg. Acc.} 

 &
\multicolumn{1}{c}{C} & \multicolumn{1}{@{}c@{}}{$\rightarrow$} & \multicolumn{1}{c}{A} & \multicolumn{1}{@{}c@{}}{$\rightarrow$} & \multicolumn{1}{c}{S} & \multicolumn{1}{@{}c@{}}{$\rightarrow$} & \multicolumn{1}{c}{P} & \multicolumn{1}{c}{Avg. Acc.} &

\multicolumn{1}{|c}{S} & \multicolumn{1}{@{}c@{}}{$\rightarrow$} & \multicolumn{1}{c}{C} & \multicolumn{1}{@{}c@{}}{$\rightarrow$} & \multicolumn{1}{c}{P} & \multicolumn{1}{@{}c@{}}{$\rightarrow$} & \multicolumn{1}{c}{A} & \multicolumn{1}{c}{Avg. Acc.} \\

\cmidrule{1-34}
\multicolumn{1}{l|}{\multirow{5}{*}{\rotatebox{90}{Non-Active}}}&$\bullet~$Source  &
N/A & \multicolumn{1}{@{}c@{}}{} & 66.3 & \multicolumn{1}{@{}c@{}}{} & 32.8 & \multicolumn{1}{@{}c@{}}{} & 43.4 & 47.5 &
N/A & \multicolumn{1}{@{}c@{}}{} & 62.8 & \multicolumn{1}{@{}c@{}}{} & 87.3 & \multicolumn{1}{@{}c@{}}{} & 55.6 & 68.5 
&
N/A & \multicolumn{1}{@{}c@{}}{} & 66.6 & \multicolumn{1}{@{}c@{}}{} & 74.4 & \multicolumn{1}{@{}c@{}}{} & 76.8 & 72.6 &
N/A & \multicolumn{1}{@{}c@{}}{} &68.6  & \multicolumn{1}{@{}c@{}}{} &50.2  & \multicolumn{1}{@{}c@{}}{} &53.4  &57.4  \\
\multicolumn{1}{l|}{}&$\bullet~$TENT~\cite{wang2020tent}  &
N/A & \multicolumn{1}{@{}c@{}}{} &70.4  & \multicolumn{1}{@{}c@{}}{} &63.3  & \multicolumn{1}{@{}c@{}}{} &59.9  &64.6  &
N/A & \multicolumn{1}{@{}c@{}}{} &73.5  & \multicolumn{1}{@{}c@{}}{} &87.2  & \multicolumn{1}{@{}c@{}}{} &56.5  &72.3  
&
N/A & \multicolumn{1}{@{}c@{}}{} &75.1  & \multicolumn{1}{@{}c@{}}{} &73.7  & \multicolumn{1}{@{}c@{}}{} &71.1  &73.3  &
N/A & \multicolumn{1}{@{}c@{}}{} &71.8  & \multicolumn{1}{@{}c@{}}{} &58.6  & \multicolumn{1}{@{}c@{}}{} &59.8  &63.4  \\
\multicolumn{1}{l|}{}&$\bullet~$COTTA~\cite{wang2022continual}  &
N/A & \multicolumn{1}{@{}c@{}}{} &67.9  & \multicolumn{1}{@{}c@{}}{} &58.4  & \multicolumn{1}{@{}c@{}}{} &72.2  &66.1  &
N/A & \multicolumn{1}{@{}c@{}}{} &73.9  & \multicolumn{1}{@{}c@{}}{} &92.3  & \multicolumn{1}{@{}c@{}}{} &69.9  &78.7 
&
N/A & \multicolumn{1}{@{}c@{}}{} &70.2  & \multicolumn{1}{@{}c@{}}{} &72.6  & \multicolumn{1}{@{}c@{}}{} &86.3  &76.4  &
N/A & \multicolumn{1}{@{}c@{}}{} &64.9  & \multicolumn{1}{@{}c@{}}{} &55.9  & \multicolumn{1}{@{}c@{}}{} &55.7  &68.8  \\
\multicolumn{1}{l|}{}&$\bullet~$SAR~\cite{niu2023towards} &
N/A & \multicolumn{1}{@{}c@{}}{} &68.3  & \multicolumn{1}{@{}c@{}}{} &63.3  & \multicolumn{1}{@{}c@{}}{} &72.7  &68.1  &
N/A & \multicolumn{1}{@{}c@{}}{} &73.6  & \multicolumn{1}{@{}c@{}}{} &93.6  & \multicolumn{1}{@{}c@{}}{} &68.7  &78.6  
&
N/A & \multicolumn{1}{@{}c@{}}{} &72.5  & \multicolumn{1}{@{}c@{}}{} &74.5  & \multicolumn{1}{@{}c@{}}{} &86.2  &77.8  &
N/A & \multicolumn{1}{@{}c@{}}{} &66.3  & \multicolumn{1}{@{}c@{}}{} &59.6  & \multicolumn{1}{@{}c@{}}{} &62.4  &62.8  \\
\multicolumn{1}{l|}{}&$\bullet~$ETA~\cite{niu2022efficient} &
N/A & \multicolumn{1}{@{}c@{}}{} &70.0  & \multicolumn{1}{@{}c@{}}{} & 64.8 & \multicolumn{1}{@{}c@{}}{} & 73.6 &69.5  &
N/A & \multicolumn{1}{@{}c@{}}{} &76.3  & \multicolumn{1}{@{}c@{}}{} &93.5  & \multicolumn{1}{@{}c@{}}{} &72.7  & 80.8 
&
N/A & \multicolumn{1}{@{}c@{}}{} &73.5  & \multicolumn{1}{@{}c@{}}{} &75.4  & \multicolumn{1}{@{}c@{}}{} &85.7  &78.2  &
N/A & \multicolumn{1}{@{}c@{}}{} &70.0  & \multicolumn{1}{@{}c@{}}{} &64.8  & \multicolumn{1}{@{}c@{}}{} &73.6  &69.5  \\
\midrule
\multicolumn{1}{l|}{\multirow{5}{*}{\rotatebox{90}{Active}}}&$\bullet~$Baseline$^{*}$  &
N/A & \multicolumn{1}{@{}c@{}}{} &68.0  & \multicolumn{1}{@{}c@{}}{} &69.9  & \multicolumn{1}{@{}c@{}}{} &70.8  &69.6  &
N/A & \multicolumn{1}{@{}c@{}}{} &78.8  & \multicolumn{1}{@{}c@{}}{} &89.3  & \multicolumn{1}{@{}c@{}}{} &67.9  &78.7  
&
N/A & \multicolumn{1}{@{}c@{}}{} &73.5  & \multicolumn{1}{@{}c@{}}{} &73.9  & \multicolumn{1}{@{}c@{}}{} &75.0  &74.2  &
N/A & \multicolumn{1}{@{}c@{}}{} &70.2  & \multicolumn{1}{@{}c@{}}{} &62.4  & \multicolumn{1}{@{}c@{}}{} &65.0  &65.9  \\
\multicolumn{1}{l|}{}&$\bullet~$SimATTA$^\dagger$ ($\mathcal{BFS}=300$)~\cite{gui2024active}  &
N/A & \multicolumn{1}{@{}c@{}}{} & 76.3 & \multicolumn{1}{@{}c@{}}{} &73.2  & \multicolumn{1}{@{}c@{}}{} &75.3  &76.3  &
N/A & \multicolumn{1}{@{}c@{}}{} &80.0  & \multicolumn{1}{@{}c@{}}{} &90.4  & \multicolumn{1}{@{}c@{}}{} &69.9  &80.1  
&
N/A & \multicolumn{1}{@{}c@{}}{} &74.5  & \multicolumn{1}{@{}c@{}}{} &75.3  & \multicolumn{1}{@{}c@{}}{} &74.3  &74.7  &
N/A & \multicolumn{1}{@{}c@{}}{} &71.6  & \multicolumn{1}{@{}c@{}}{} &73.4  & \multicolumn{1}{@{}c@{}}{} &68.2  &71.1 \\

% \multicolumn{1}{l|}{}&\cellcolor{mycolor1}$\bullet~$Ours$^{*}$  &\cellcolor{mycolor1}
% N/A &\cellcolor{mycolor1} \multicolumn{1}{@{}c@{}}{}  &\cellcolor{mycolor1}74.9  &\cellcolor{mycolor1} \multicolumn{1}{@{}c@{}}{} &\cellcolor{mycolor1}74.5  &\cellcolor{mycolor1} \multicolumn{1}{@{}c@{}}{} &\cellcolor{mycolor1}76.7  &\cellcolor{mycolor1}75.3  &\cellcolor{mycolor1}
% N/A & \cellcolor{mycolor1} \multicolumn{1}{@{}c@{}}{} &\cellcolor{mycolor1}81.1  & \cellcolor{mycolor1} \multicolumn{1}{@{}c@{}}{} &\cellcolor{mycolor1}91.6  & \cellcolor{mycolor1} \multicolumn{1}{@{}c@{}}{} &\cellcolor{mycolor1}75.6  &\cellcolor{mycolor1}82.8  
% &\cellcolor{mycolor1} N/A &\cellcolor{mycolor1} \multicolumn{1}{@{}c@{}}{} &\cellcolor{mycolor1}77.0  &\cellcolor{mycolor1} \multicolumn{1}{@{}c@{}}{} &\cellcolor{mycolor1}78.2  &\cellcolor{mycolor1} \multicolumn{1}{@{}c@{}}{} &\cellcolor{mycolor1}81.9  &\cellcolor{mycolor1}79.0  &\cellcolor{mycolor1}
% N/A & \cellcolor{mycolor1} \multicolumn{1}{@{}c@{}}{} &\cellcolor{mycolor1}75.2  & \cellcolor{mycolor1} \multicolumn{1}{@{}c@{}}{} &\cellcolor{mycolor1}62.9  &\cellcolor{mycolor1} \multicolumn{1}{@{}c@{}}{} &\cellcolor{mycolor1}68.8  &\cellcolor{mycolor1}69.0  

\multicolumn{1}{l|}{}&
\multicolumn{1}{l|}{\cellcolor{mycolor1}$\bullet~$Ours$^{*}$} &
\multicolumn{1}{c|}{\cellcolor{mycolor1}N/A} & 
\multicolumn{1}{@{}c@{}}{\cellcolor{mycolor1}} & 
\multicolumn{1}{c|}{\cellcolor{mycolor1}74.9} & 
\multicolumn{1}{@{}c@{}}{\cellcolor{mycolor1}} & 
\multicolumn{1}{c|}{\cellcolor{mycolor1}74.5} & 
\multicolumn{1}{@{}c@{}}{\cellcolor{mycolor1}} & 
\multicolumn{1}{c}{\cellcolor{mycolor1}76.7} & 
\multicolumn{1}{c|}{\cellcolor{mycolor1}75.3} & 
\multicolumn{1}{c|}{\cellcolor{mycolor1}N/A} & 
\multicolumn{1}{@{}c@{}}{\cellcolor{mycolor1}} & 
\multicolumn{1}{c|}{\cellcolor{mycolor1}81.1} & 
\multicolumn{1}{@{}c@{}}{\cellcolor{mycolor1}} & 
\multicolumn{1}{c|}{\cellcolor{mycolor1}91.6} & 
\multicolumn{1}{@{}c@{}}{\cellcolor{mycolor1}} & 
\multicolumn{1}{c}{\cellcolor{mycolor1}75.6} & 
\multicolumn{1}{c|}{\cellcolor{mycolor1}82.8} & 
\multicolumn{1}{c|}{\cellcolor{mycolor1}N/A} & 
\multicolumn{1}{@{}c@{}}{\cellcolor{mycolor1}} & 
\multicolumn{1}{c|}{\cellcolor{mycolor1}77.0} & 
\multicolumn{1}{@{}c@{}}{\cellcolor{mycolor1}} & 
\multicolumn{1}{c|}{\cellcolor{mycolor1}78.2} & 
\multicolumn{1}{@{}c@{}}{\cellcolor{mycolor1}} & 
\multicolumn{1}{c}{\cellcolor{mycolor1}81.9} & 
\multicolumn{1}{c}{\cellcolor{mycolor1}79.0} & 
\multicolumn{1}{c|}{\cellcolor{mycolor1}N/A} & 
\multicolumn{1}{@{}c@{}}{\cellcolor{mycolor1}} & 
\multicolumn{1}{c|}{\cellcolor{mycolor1}75.2} & 
\multicolumn{1}{@{}c@{}}{\cellcolor{mycolor1}} & 
\multicolumn{1}{c|}{\cellcolor{mycolor1}62.9} & 
\multicolumn{1}{@{}c@{}}{\cellcolor{mycolor1}} & 
\multicolumn{1}{c}{\cellcolor{mycolor1}68.8} & 
\multicolumn{1}{c}{\cellcolor{mycolor1}69.0} \\

% \multicolumn{1}{l|}{}&\cellcolor{mycolor2}$\bullet~$Ours$^\dagger$  &\cellcolor{mycolor2}
% N/A & \cellcolor{mycolor2} \multicolumn{1}{@{}c@{}}{} & \cellcolor{mycolor2}\underline{76.7} &\cellcolor{mycolor2} \multicolumn{1}{@{}c@{}}{} &\cellcolor{mycolor2}\underline{78.4}  & \cellcolor{mycolor2} \multicolumn{1}{@{}c@{}}{} &\cellcolor{mycolor2}\underline{78.8}  &\cellcolor{mycolor2}\underline{77.9}  &\cellcolor{mycolor2}
% N/A &\cellcolor{mycolor2} \multicolumn{1}{@{}c@{}}{} &\cellcolor{mycolor2}\textbf{82.9} & \cellcolor{mycolor2} \multicolumn{1}{@{}c@{}}{} &\cellcolor{mycolor2}\underline{90.1}  & \cellcolor{mycolor2} \multicolumn{1}{@{}c@{}}{} &\cellcolor{mycolor2}\underline{77.2}  &\cellcolor{mycolor2}\underline{83.4} 
% &\cellcolor{mycolor2}
% N/A &\cellcolor{mycolor2} \multicolumn{1}{@{}c@{}}{} &\cellcolor{mycolor2}\underline{78.1}  &\cellcolor{mycolor2} \multicolumn{1}{@{}c@{}}{} &\cellcolor{mycolor2}\textbf{79.5}  &\cellcolor{mycolor2}  \multicolumn{1}{@{}c@{}}{} &\cellcolor{mycolor2}\underline{83.0}  &\cellcolor{mycolor2}\underline{80.2}  &\cellcolor{mycolor2}
% N/A & \cellcolor{mycolor2} \multicolumn{1}{@{}c@{}}{} &\cellcolor{mycolor2}\underline{78.4}  &\cellcolor{mycolor2} \multicolumn{1}{@{}c@{}}{} &\cellcolor{mycolor2}\underline{86.9}  &\cellcolor{mycolor2} \multicolumn{1}{@{}c@{}}{} &\cellcolor{mycolor2}\underline{75.7}  &\cellcolor{mycolor2}\underline{80.3}  \\
\multicolumn{1}{l|}{}&
\multicolumn{1}{l|}{\cellcolor{mycolor2}$\bullet~$Ours$^\dagger$} &
\multicolumn{1}{c|}{\cellcolor{mycolor2}N/A} & 
\multicolumn{1}{@{}c@{}}{\cellcolor{mycolor2}} & 
\multicolumn{1}{c|}{\cellcolor{mycolor2}\underline{76.7}} & 
\multicolumn{1}{@{}c@{}}{\cellcolor{mycolor2}} & 
\multicolumn{1}{c|}{\cellcolor{mycolor2}\underline{78.4}} & 
\multicolumn{1}{@{}c@{}}{\cellcolor{mycolor2}} & 
\multicolumn{1}{c}{\cellcolor{mycolor2}\underline{78.8}} & 
\multicolumn{1}{c|}{\cellcolor{mycolor2}\underline{77.9}} & 
\multicolumn{1}{c|}{\cellcolor{mycolor2}N/A} & 
\multicolumn{1}{@{}c@{}}{\cellcolor{mycolor2}} & 
\multicolumn{1}{c|}{\cellcolor{mycolor2}\textbf{82.9}} & 
\multicolumn{1}{@{}c@{}}{\cellcolor{mycolor2}} & 
\multicolumn{1}{c|}{\cellcolor{mycolor2}\underline{90.1}} & 
\multicolumn{1}{@{}c@{}}{\cellcolor{mycolor2}} & 
\multicolumn{1}{c}{\cellcolor{mycolor2}\underline{77.2}} & 
\multicolumn{1}{c|}{\cellcolor{mycolor2}\underline{83.4}} & 
\multicolumn{1}{c|}{\cellcolor{mycolor2}N/A} & 
\multicolumn{1}{@{}c@{}}{\cellcolor{mycolor2}} & 
\multicolumn{1}{c|}{\cellcolor{mycolor2}\underline{78.1}} & 
\multicolumn{1}{@{}c@{}}{\cellcolor{mycolor2}} & 
\multicolumn{1}{c|}{\cellcolor{mycolor2}\textbf{79.5}} & 
\multicolumn{1}{@{}c@{}}{\cellcolor{mycolor2}} & 
\multicolumn{1}{c}{\cellcolor{mycolor2}\underline{83.0}} & 
\multicolumn{1}{c|}{\cellcolor{mycolor2}\underline{80.2}} & 
\multicolumn{1}{c|}{\cellcolor{mycolor2}N/A} & 
\multicolumn{1}{@{}c@{}}{\cellcolor{mycolor2}} & 
\multicolumn{1}{c|}{\cellcolor{mycolor2}\underline{78.4}} & 
\multicolumn{1}{@{}c@{}}{\cellcolor{mycolor2}} & 
\multicolumn{1}{c|}{\cellcolor{mycolor2}\underline{86.9}} & 
\multicolumn{1}{@{}c@{}}{\cellcolor{mycolor2}} & 
\multicolumn{1}{c}{\cellcolor{mycolor2}\underline{75.7}} & 
\multicolumn{1}{c}{\cellcolor{mycolor2}\underline{80.3}} \\

% \multicolumn{1}{l|}{}&\cellcolor{mycolor3}$\bullet~$Ours$^\dagger$ ($\mathcal{BFS}=300$)  &\cellcolor{mycolor3}
% N/A & \cellcolor{mycolor3} \multicolumn{1}{@{}c@{}}{} &\cellcolor{mycolor3}\textbf{78.2}  &\cellcolor{mycolor3} \multicolumn{1}{@{}c@{}}{} &\cellcolor{mycolor3}\textbf{79.4}  &\cellcolor{mycolor3} 
%  \multicolumn{1}{@{}c@{}}{} &\cellcolor{mycolor3}\textbf{79.7}  &\cellcolor{mycolor3}\textbf{79.1}  &\cellcolor{mycolor3}
% N/A &\cellcolor{mycolor3} \multicolumn{1}{@{}c@{}}{} &\cellcolor{mycolor3}\underline{82.8}  &\cellcolor{mycolor3} \multicolumn{1}{@{}c@{}}{} &\cellcolor{mycolor3}\textbf{92.3}  & \cellcolor{mycolor3} \multicolumn{1}{@{}c@{}}{} &\cellcolor{mycolor3}\textbf{80.6}  &\cellcolor{mycolor3}\textbf{85.2}  
% &\cellcolor{mycolor3}
% N/A &\cellcolor{mycolor3} \multicolumn{1}{@{}c@{}}{} &\cellcolor{mycolor3}\textbf{78.2}  &\cellcolor{mycolor3} \multicolumn{1}{@{}c@{}}{} &\cellcolor{mycolor3}\underline{79.1} & \cellcolor{mycolor3} \multicolumn{1}{@{}c@{}}{} &\cellcolor{mycolor3}\textbf{86.0}  &\cellcolor{mycolor3}\textbf{81.1}  &\cellcolor{mycolor3}
% N/A &\cellcolor{mycolor3} \multicolumn{1}{@{}c@{}}{} &\cellcolor{mycolor3}\textbf{79.7}  &\cellcolor{mycolor3} \multicolumn{1}{@{}c@{}}{} &\cellcolor{mycolor3}\textbf{87.5}  &\cellcolor{mycolor3} \multicolumn{1}{@{}c@{}}{} &\cellcolor{mycolor3}\textbf{79.9}  &\cellcolor{mycolor3}\textbf{82.4} \\
\multicolumn{1}{l|}{}&
\multicolumn{1}{l|}{\cellcolor{mycolor3}$\bullet~$Ours$^\dagger$ ($\mathcal{BFS}=300$)} &
\multicolumn{1}{c|}{\cellcolor{mycolor3}N/A} & 
\multicolumn{1}{@{}c@{}}{\cellcolor{mycolor3}} & 
\multicolumn{1}{c|}{\cellcolor{mycolor3}\textbf{78.2}} & 
\multicolumn{1}{@{}c@{}}{\cellcolor{mycolor3}} & 
\multicolumn{1}{c|}{\cellcolor{mycolor3}\textbf{79.4}} & 
\multicolumn{1}{@{}c@{}}{\cellcolor{mycolor3}} & 
\multicolumn{1}{c}{\cellcolor{mycolor3}\textbf{79.7}} & 
\multicolumn{1}{c|}{\cellcolor{mycolor3}\textbf{79.1}} & 
\multicolumn{1}{c|}{\cellcolor{mycolor3}N/A} & 
\multicolumn{1}{@{}c@{}}{\cellcolor{mycolor3}} & 
\multicolumn{1}{c|}{\cellcolor{mycolor3}\underline{82.8}} & 
\multicolumn{1}{@{}c@{}}{\cellcolor{mycolor3}} & 
\multicolumn{1}{c|}{\cellcolor{mycolor3}\textbf{92.3}} & 
\multicolumn{1}{@{}c@{}}{\cellcolor{mycolor3}} & 
\multicolumn{1}{c}{\cellcolor{mycolor3}\textbf{80.6}} & 
\multicolumn{1}{c|}{\cellcolor{mycolor3}\textbf{85.2}} & 
\multicolumn{1}{c|}{\cellcolor{mycolor3}N/A} & 
\multicolumn{1}{@{}c@{}}{\cellcolor{mycolor3}} & 
\multicolumn{1}{c|}{\cellcolor{mycolor3}\textbf{78.2}} & 
\multicolumn{1}{@{}c@{}}{\cellcolor{mycolor3}} & 
\multicolumn{1}{c|}{\cellcolor{mycolor3}\underline{79.1}} & 
\multicolumn{1}{@{}c@{}}{\cellcolor{mycolor3}} & 
\multicolumn{1}{c}{\cellcolor{mycolor3}\textbf{86.0}} & 
\multicolumn{1}{c|}{\cellcolor{mycolor3}\textbf{81.1}} & 
\multicolumn{1}{c|}{\cellcolor{mycolor3}N/A} & 
\multicolumn{1}{@{}c@{}}{\cellcolor{mycolor3}} & 
\multicolumn{1}{c|}{\cellcolor{mycolor3}\textbf{79.7}} & 
\multicolumn{1}{@{}c@{}}{\cellcolor{mycolor3}} & 
\multicolumn{1}{c|}{\cellcolor{mycolor3}\textbf{87.5}} & 
\multicolumn{1}{@{}c@{}}{\cellcolor{mycolor3}} & 
\multicolumn{1}{c}{\cellcolor{mycolor3}\textbf{79.9}} & 
\multicolumn{1}{c}{\cellcolor{mycolor3}\textbf{82.4}} \\

\bottomrule

\end{tabular}
}
\end{center}

\begin{center}
 \vspace{-0.1in}
 \resizebox{1.0\linewidth}{!}{
 
\begin{tabular}{ll|cccccccc|cccccccc|cccccccc|cccccccc}
\toprule

\multicolumn{1}{l|}{\textbf{F}}&\multicolumn{1}{l|}{Methods} &
\multicolumn{1}{c}{P} & \multicolumn{1}{@{}c@{}}{$\rightarrow$} & \multicolumn{1}{c}{A} & \multicolumn{1}{@{}c@{}}{$\rightarrow$} & \multicolumn{1}{c}{C} & \multicolumn{1}{@{}c@{}}{$\rightarrow$} & \multicolumn{1}{c}{S} & \multicolumn{1}{c}{Avg. Acc.} &

\multicolumn{1}{|c}{A} & \multicolumn{1}{@{}c@{}}{$\rightarrow$} & \multicolumn{1}{c}{C} & \multicolumn{1}{@{}c@{}}{$\rightarrow$} & \multicolumn{1}{c}{P} & \multicolumn{1}{@{}c@{}}{$\rightarrow$} & \multicolumn{1}{c}{S} & \multicolumn{1}{c|}{Avg. Acc.} 

 &
\multicolumn{1}{c}{C} & \multicolumn{1}{@{}c@{}}{$\rightarrow$} & \multicolumn{1}{c}{A} & \multicolumn{1}{@{}c@{}}{$\rightarrow$} & \multicolumn{1}{c}{S} & \multicolumn{1}{@{}c@{}}{$\rightarrow$} & \multicolumn{1}{c}{P} & \multicolumn{1}{c}{Avg. Acc.} &

\multicolumn{1}{|c}{S} & \multicolumn{1}{@{}c@{}}{$\rightarrow$} & \multicolumn{1}{c}{C} & \multicolumn{1}{@{}c@{}}{$\rightarrow$} & \multicolumn{1}{c}{P} & \multicolumn{1}{@{}c@{}}{$\rightarrow$} & \multicolumn{1}{c}{A} & \multicolumn{1}{c}{Avg. Acc.} \\

\midrule
\multicolumn{1}{l|}{\multirow{5}{*}{\rotatebox{90}{Non-Active}}}&$\bullet~$Source  &
N/A & \multicolumn{1}{@{}c@{}}{} & 66.3 & \multicolumn{1}{@{}c@{}}{} & 32.8 & \multicolumn{1}{@{}c@{}}{} & 43.4 & 47.5 &
N/A & \multicolumn{1}{@{}c@{}}{} & 62.8 & \multicolumn{1}{@{}c@{}}{} & 87.3 & \multicolumn{1}{@{}c@{}}{} & 55.6 & 68.5 
&
N/A & \multicolumn{1}{@{}c@{}}{} & 66.6 & \multicolumn{1}{@{}c@{}}{} & 74.4 & \multicolumn{1}{@{}c@{}}{} & 76.8 & 72.6 &
N/A & \multicolumn{1}{@{}c@{}}{} &68.6  & \multicolumn{1}{@{}c@{}}{} &50.2  & \multicolumn{1}{@{}c@{}}{} &53.4  &57.4  \\
\multicolumn{1}{l|}{}&$\bullet~$TENT~\cite{wang2020tent}  &
N/A & \multicolumn{1}{@{}c@{}}{} &70.4  & \multicolumn{1}{@{}c@{}}{} &64.3  & \multicolumn{1}{@{}c@{}}{} &75.4  &70.1  &
N/A & \multicolumn{1}{@{}c@{}}{} &73.5  & \multicolumn{1}{@{}c@{}}{} &94.9  & \multicolumn{1}{@{}c@{}}{} &68.3  & 78.9 
&
N/A & \multicolumn{1}{@{}c@{}}{} &75.1  & \multicolumn{1}{@{}c@{}}{} &76.8  & \multicolumn{1}{@{}c@{}}{} &86.2  &79.4  &
N/A & \multicolumn{1}{@{}c@{}}{} &71.8  & \multicolumn{1}{@{}c@{}}{} &60.2  & \multicolumn{1}{@{}c@{}}{} &63.7  &65.2 \\
\multicolumn{1}{l|}{}&$\bullet~$COTTA~\cite{wang2022continual}  &
N/A & \multicolumn{1}{@{}c@{}}{} &67.9  & \multicolumn{1}{@{}c@{}}{} & 58.1 & \multicolumn{1}{@{}c@{}}{} &72.3  & 66.1 &
N/A & \multicolumn{1}{@{}c@{}}{} &73.9  & \multicolumn{1}{@{}c@{}}{} &92.5  & \multicolumn{1}{@{}c@{}}{} &68.9  & 78.4 
&
N/A & \multicolumn{1}{@{}c@{}}{} &70.2  & \multicolumn{1}{@{}c@{}}{} &72.8  & \multicolumn{1}{@{}c@{}}{} &85.3  &76.1  &
N/A & \multicolumn{1}{@{}c@{}}{} &64.9  & \multicolumn{1}{@{}c@{}}{} &55.3  & \multicolumn{1}{@{}c@{}}{} &54.9  &58.4  \\
\multicolumn{1}{l|}{}&$\bullet~$SAR~\cite{niu2023towards}  &
N/A & \multicolumn{1}{@{}c@{}}{} &68.3  & \multicolumn{1}{@{}c@{}}{} &60.9  & \multicolumn{1}{@{}c@{}}{} &72.6  &67.2  &
N/A & \multicolumn{1}{@{}c@{}}{} &73.6  & \multicolumn{1}{@{}c@{}}{} &93.1  & \multicolumn{1}{@{}c@{}}{} &71.4  &79.4
&
N/A & \multicolumn{1}{@{}c@{}}{} &72.5  & \multicolumn{1}{@{}c@{}}{} &74.9  & \multicolumn{1}{@{}c@{}}{} &86.0  &77.8  &
N/A & \multicolumn{1}{@{}c@{}}{} &66.2  & \multicolumn{1}{@{}c@{}}{} &59.6  & \multicolumn{1}{@{}c@{}}{} &60.1  &62.0  \\
\multicolumn{1}{l|}{}&$\bullet~$ETA~\cite{niu2022efficient}  &
N/A & \multicolumn{1}{@{}c@{}}{} &70.0  & \multicolumn{1}{@{}c@{}}{} &63.3  & \multicolumn{1}{@{}c@{}}{} &73.7  &69.0  &
N/A & \multicolumn{1}{@{}c@{}}{} &76.3  & \multicolumn{1}{@{}c@{}}{} & 94.7 & \multicolumn{1}{@{}c@{}}{} &73.8  & 81.6
&
N/A & \multicolumn{1}{@{}c@{}}{} &73.5  & \multicolumn{1}{@{}c@{}}{} &76.5  & \multicolumn{1}{@{}c@{}}{} &87.0  &79.0  &
N/A & \multicolumn{1}{@{}c@{}}{} &70.0  & \multicolumn{1}{@{}c@{}}{} &63.3  & \multicolumn{1}{@{}c@{}}{} & 73.7 &69.0  \\
\cmidrule{1-34}
\multicolumn{1}{l|}{\multirow{5}{*}{\rotatebox{90}{Active}}}&$\bullet~$Baseline$^{*}$    &
N/A & \multicolumn{1}{@{}c@{}}{} &68.0  & \multicolumn{1}{@{}c@{}}{} &68.2  & \multicolumn{1}{@{}c@{}}{} &73.8  &70.0  &
N/A & \multicolumn{1}{@{}c@{}}{} &78.8  & \multicolumn{1}{@{}c@{}}{} &94.7  & \multicolumn{1}{@{}c@{}}{} &70.5  &81.3  
&
N/A & \multicolumn{1}{@{}c@{}}{} &73.5  & \multicolumn{1}{@{}c@{}}{} &74.7  & \multicolumn{1}{@{}c@{}}{} &85.7  &78.0  &
N/A & \multicolumn{1}{@{}c@{}}{} &70.2  & \multicolumn{1}{@{}c@{}}{} &64.8  & \multicolumn{1}{@{}c@{}}{} & 62.3 &65.9  \\

\multicolumn{1}{l|}{}&$\bullet~$SimATTA$^\dagger$ ($\mathcal{BFS}=300$)~\cite{gui2024active}  &
N/A & \multicolumn{1}{@{}c@{}}{} &76.3  & \multicolumn{1}{@{}c@{}}{} &69.5  & \multicolumn{1}{@{}c@{}}{} &75.5  &73.8  &
N/A & \multicolumn{1}{@{}c@{}}{} &80.0  & \multicolumn{1}{@{}c@{}}{} &93.0  & \multicolumn{1}{@{}c@{}}{} &71.9  &81.6  
&
N/A & \multicolumn{1}{@{}c@{}}{} &74.5  & \multicolumn{1}{@{}c@{}}{} &76.9  & \multicolumn{1}{@{}c@{}}{} &87.6  &79.7  &
N/A & \multicolumn{1}{@{}c@{}}{} &71.6  & \multicolumn{1}{@{}c@{}}{} &73.3  & \multicolumn{1}{@{}c@{}}{} &66.3  &70.4 \\

\multicolumn{1}{l|}{}&
\multicolumn{1}{l|}{\cellcolor{mycolor1}$\bullet~$Ours$^{*}$} &
\multicolumn{1}{c|}{\cellcolor{mycolor1}N/A} & 
\multicolumn{1}{@{}c@{}}{\cellcolor{mycolor1}} & 
\multicolumn{1}{c|}{\cellcolor{mycolor1}74.9} & 
\multicolumn{1}{@{}c@{}}{\cellcolor{mycolor1}} & 
\multicolumn{1}{c|}{\cellcolor{mycolor1}70.8} & 
\multicolumn{1}{@{}c@{}}{\cellcolor{mycolor1}} & 
\multicolumn{1}{c}{\cellcolor{mycolor1}75.4} & 
\multicolumn{1}{c|}{\cellcolor{mycolor1}73.7} & 
\multicolumn{1}{c|}{\cellcolor{mycolor1}N/A} & 
\multicolumn{1}{@{}c@{}}{\cellcolor{mycolor1}} & 
\multicolumn{1}{c|}{\cellcolor{mycolor1}81.1} & 
\multicolumn{1}{@{}c@{}}{\cellcolor{mycolor1}} & 
\multicolumn{1}{c|}{\cellcolor{mycolor1}95.3} & 
\multicolumn{1}{@{}c@{}}{\cellcolor{mycolor1}} & 
\multicolumn{1}{c}{\cellcolor{mycolor1}81.7} & 
\multicolumn{1}{c|}{\cellcolor{mycolor1}86.0} & 
\multicolumn{1}{c|}{\cellcolor{mycolor1}N/A} & 
\multicolumn{1}{@{}c@{}}{\cellcolor{mycolor1}} & 
\multicolumn{1}{c|}{\cellcolor{mycolor1}77.0} & 
\multicolumn{1}{@{}c@{}}{\cellcolor{mycolor1}} & 
\multicolumn{1}{c|}{\cellcolor{mycolor1}77.0} & 
\multicolumn{1}{@{}c@{}}{\cellcolor{mycolor1}} & 
\multicolumn{1}{c}{\cellcolor{mycolor1}\underline{88.7}} & 
\multicolumn{1}{c|}{\cellcolor{mycolor1}80.9} & 
\multicolumn{1}{c|}{\cellcolor{mycolor1}N/A} & 
\multicolumn{1}{@{}c@{}}{\cellcolor{mycolor1}} & 
\multicolumn{1}{c|}{\cellcolor{mycolor1}75.2} & 
\multicolumn{1}{@{}c@{}}{\cellcolor{mycolor1}} & 
\multicolumn{1}{c|}{\cellcolor{mycolor1}62.6} & 
\multicolumn{1}{@{}c@{}}{\cellcolor{mycolor1}} & 
\multicolumn{1}{c}{\cellcolor{mycolor1}65.8} & 
\multicolumn{1}{c}{\cellcolor{mycolor1}67.9} \\

\multicolumn{1}{l|}{}&
\multicolumn{1}{l|}{\cellcolor{mycolor2}$\bullet~$Ours$^\dagger$} &
\multicolumn{1}{c|}{\cellcolor{mycolor2}N/A} & 
\multicolumn{1}{@{}c@{}}{\cellcolor{mycolor2}} & 
\multicolumn{1}{c|}{\cellcolor{mycolor2}\underline{76.7}} & 
\multicolumn{1}{@{}c@{}}{\cellcolor{mycolor2}} & 
\multicolumn{1}{c|}{\cellcolor{mycolor2}\textbf{74.4}} & 
\multicolumn{1}{@{}c@{}}{\cellcolor{mycolor2}} & 
\multicolumn{1}{c}{\cellcolor{mycolor2}\underline{75.7}} & 
\multicolumn{1}{c|}{\cellcolor{mycolor2}\underline{75.6}} & 
\multicolumn{1}{c|}{\cellcolor{mycolor2}N/A} & 
\multicolumn{1}{@{}c@{}}{\cellcolor{mycolor2}} & 
\multicolumn{1}{c|}{\cellcolor{mycolor2}\textbf{82.9}} & 
\multicolumn{1}{@{}c@{}}{\cellcolor{mycolor2}} & 
\multicolumn{1}{c|}{\cellcolor{mycolor2}\underline{95.4}} & 
\multicolumn{1}{@{}c@{}}{\cellcolor{mycolor2}} & 
\multicolumn{1}{c}{\cellcolor{mycolor2}\underline{82.1}} & 
\multicolumn{1}{c|}{\cellcolor{mycolor2}\underline{86.8}} & 
\multicolumn{1}{c|}{\cellcolor{mycolor2}N/A} & 
\multicolumn{1}{@{}c@{}}{\cellcolor{mycolor2}} & 
\multicolumn{1}{c|}{\cellcolor{mycolor2}\underline{78.1}} & 
\multicolumn{1}{@{}c@{}}{\cellcolor{mycolor2}} & 
\multicolumn{1}{c|}{\cellcolor{mycolor2}\underline{80.1}} & 
\multicolumn{1}{@{}c@{}}{\cellcolor{mycolor2}} & 
\multicolumn{1}{c}{\cellcolor{mycolor2}\underline{88.7}} & 
\multicolumn{1}{c|}{\cellcolor{mycolor2}\underline{82.3}} & 
\multicolumn{1}{c|}{\cellcolor{mycolor2}N/A} & 
\multicolumn{1}{@{}c@{}}{\cellcolor{mycolor2}} & 
\multicolumn{1}{c|}{\cellcolor{mycolor2}\underline{78.4}} & 
\multicolumn{1}{@{}c@{}}{\cellcolor{mycolor2}} & 
\multicolumn{1}{c|}{\cellcolor{mycolor2}\textbf{81.1}} & 
\multicolumn{1}{@{}c@{}}{\cellcolor{mycolor2}} & 
\multicolumn{1}{c}{\cellcolor{mycolor2}\underline{68.9}} & 
\multicolumn{1}{c}{\cellcolor{mycolor2}\underline{76.1}} \\

\multicolumn{1}{l|}{}&
\multicolumn{1}{l|}{\cellcolor{mycolor3}$\bullet~$Ours$^\dagger$ ($\mathcal{BFS}=300$)} &
\multicolumn{1}{c|}{\cellcolor{mycolor3}N/A} & 
\multicolumn{1}{@{}c@{}}{\cellcolor{mycolor3}} & 
\multicolumn{1}{c|}{\cellcolor{mycolor3}\textbf{78.2}} & 
\multicolumn{1}{@{}c@{}}{\cellcolor{mycolor3}} & 
\multicolumn{1}{c|}{\cellcolor{mycolor3}\underline{74.3}} & 
\multicolumn{1}{@{}c@{}}{\cellcolor{mycolor3}} & 
\multicolumn{1}{c}{\cellcolor{mycolor3}\textbf{80.6}} & 
\multicolumn{1}{c|}{\cellcolor{mycolor3}\textbf{77.7}} & 
\multicolumn{1}{c|}{\cellcolor{mycolor3}N/A} & 
\multicolumn{1}{@{}c@{}}{\cellcolor{mycolor3}} & 
\multicolumn{1}{c|}{\cellcolor{mycolor3}\underline{82.7}} & 
\multicolumn{1}{@{}c@{}}{\cellcolor{mycolor3}} & 
\multicolumn{1}{c|}{\cellcolor{mycolor3}\textbf{95.7}} & 
\multicolumn{1}{@{}c@{}}{\cellcolor{mycolor3}} & 
\multicolumn{1}{c}{\cellcolor{mycolor3}\textbf{82.5}} & 
\multicolumn{1}{c|}{\cellcolor{mycolor3}\textbf{87.0}} & 
\multicolumn{1}{c|}{\cellcolor{mycolor3}N/A} & 
\multicolumn{1}{@{}c@{}}{\cellcolor{mycolor3}} & 
\multicolumn{1}{c|}{\cellcolor{mycolor3}\textbf{78.2}} & 
\multicolumn{1}{@{}c@{}}{\cellcolor{mycolor3}} & 
\multicolumn{1}{c|}{\cellcolor{mycolor3}\textbf{81.3}} & 
\multicolumn{1}{@{}c@{}}{\cellcolor{mycolor3}} & 
\multicolumn{1}{c}{\cellcolor{mycolor3}\textbf{90.4}} & 
\multicolumn{1}{c|}{\cellcolor{mycolor3}\textbf{83.3}} & 
\multicolumn{1}{c|}{\cellcolor{mycolor3}N/A} & 
\multicolumn{1}{@{}c@{}}{\cellcolor{mycolor3}} & 
\multicolumn{1}{c|}{\cellcolor{mycolor3}\textbf{79.7}} & 
\multicolumn{1}{@{}c@{}}{\cellcolor{mycolor3}} & 
\multicolumn{1}{c|}{\cellcolor{mycolor3}\underline{77.2}} & 
\multicolumn{1}{@{}c@{}}{\cellcolor{mycolor3}} & 
\multicolumn{1}{c}{\cellcolor{mycolor3}\textbf{74.2}} & 
\multicolumn{1}{c}{\cellcolor{mycolor3}\textbf{77.0}} \\

\bottomrule

\end{tabular}}
\end{center}

\end{table*}
\noindent learning rates on ImageNet-C, ImageNet-R/K, and ImageNet-A are set to 0.00025, 0.001, and 0.005, respectively. We adopt the SGD optimizer with a momentum of 0.9 on the above databases. For PACS , we use the Adam optimizer with a learning rate of 0.005. We run all experiments on an NVIDIA RTX 4090D GPU.

\subsection{Robustness to Synthetic Corruptions} 
As shown in Table~\ref{tab:imagenet-c}, compared to the TENT approach~\cite{wang2020tent}, the ATTA baseline reduces the average error rate from 70.9\% to 60.6\% in the CTTA setting. This result highlights the importance of involving active learning for long-term TTA. 
Our approach outperforms the baseline by 6.8\% with one oracle label per batch.
Additionally, it surpasses SimATTA~\cite{gui2024active} and HILTTA~\cite{li2024exploring} by 3.9\% and 3.2\% in terms of average error rate, respectively, using the same annotation count and buffer settings. More impressively, the basic version of our approach (\textit{i.e.}, Ours* in Table~\ref{tab:imagenet-c}) outperforms HILTTA that requires triple-times annotations by 1.3\%. It also outperforms SimATTA by 1.0\% that requires both triple-times annotations and a special buffer to store annotated samples. It is worth noting that both more annotations and the special buffer reduce ATTA efficiency.

SimATTA and CEMA~\cite{chen2024towards} are specifically designed for the FTTA setting and they achieve promising results in Table~\ref{tab:imagenet-c}. Our method outperforms both SimATTA and CEMA in the FTTA setting. In particular, it outperforms SimATTA by 3.4\% in terms of the average error rate when they adopt the same annotation number and buffer settings. Moreover, the basic version of our approach (\textit{i.e.}, Ours* in Table~\ref{tab:imagenet-c}) still outperforms SimATTA that requires triple-times annotations and a special buffer by 1.7\%.

\subsection{Robustness to Real-World Distribution Shifts}
We further make comparisons on ImageNet-R~\cite{hendrycks2021many}, ImageNet-K~\cite{wang2019learning}, ImageNet-A~\cite{hendrycks2021nae}, and PACS~\cite{li2017deeper}. The experimental results are summarized in Table~\ref{tab:imagenet-variants} and Table~\ref{tab:pacs}. 
In Table~\ref{tab:imagenet-variants}, it is shown that our method continues to achieve the best performance. Compared with the ATTA baseline, it reduces the average error rates by 5.1\%, 3.1\%, and 2.5\% on ImageNet-R, ImageNet-K, and ImageNet-A, respectively. Moreover, under the same annotation number and buffer settings, our method significantly outperforms SimATTA~\cite{gui2024active} in terms of the average error rate by 3.2\%, 2.0\%, and 3.1\% on ImageNet-R, ImageNet-K, and ImageNet-A, respectively.

For experiments on PACS, we evaluate the performance of different TTA methods across four different sequences of domain changes, as shown in Table~\ref{tab:pacs}.
\begin{table}[htp]
\caption{Performance comparisons on the key components of EATTA. All settings in this table annotate only one sample per batch.}
\label{tab:components}
\vspace{-0.1in}
\resizebox{1.0\linewidth}{!}{
\begin{tabular}{l|cccc|ccccc}
\toprule
 & PD & CB   &GND &EMA     &ImageNet-C  &ImageNet-R  &ImageNet-K  &ImageNet-A &Avg. Err. \\  \toprule
    
\ding{172}& \multicolumn{1}{c}{\ding{55}}  &\ding{55}   &\ding{55}  & \multicolumn{1}{c|}{\ding{55}} &59.6  & 54.9 &66.9  &99.4 &70.2 \\

\ding{173}&\multicolumn{1}{c}{\ding{51}}  &\ding{55}   &\ding{55}  & \multicolumn{1}{c|}{\ding{55}} &58.6  &53.3  &65.7  &99.2 &69.2 \\

\ding{174}&\multicolumn{1}{c}{\ding{51}}  &\ding{51}   &\ding{55}  & \multicolumn{1}{c|}{\ding{55}} &57.3  &52.6  &65.5  &99.0 &68.6 \\

\ding{175}&\multicolumn{1}{c}{\ding{51}}  &\ding{55}   &\ding{51}  & \multicolumn{1}{c|}{\ding{55}} &55.2  &52.0  &65.3  &98.7 &67.8 \\

\ding{176}&\multicolumn{1}{c}{\ding{51}}  &\ding{55}   &\ding{51}  & \multicolumn{1}{c|}{\ding{51}} &\underline{54.3}  &\underline{51.7}  &\underline{64.9}  &98.4 &\underline{67.3} \\

\ding{177}&\multicolumn{1}{c}{\ding{55}}  &\ding{55}   &\ding{51}  & \multicolumn{1}{c|}{\ding{55}} &55.6  &54.4  &65.6  &98.7 &68.6 \\

\ding{178}&\multicolumn{1}{c}{\ding{55}}  &\ding{55}   &\ding{51}  & \multicolumn{1}{c|}{\ding{51}} &54.7  &53.5  &65.4  &98.5 &68.0 \\

\ding{179}&\multicolumn{1}{c}{\ding{51}}  &\ding{51}   &\ding{51}  & \multicolumn{1}{c|}{\ding{55}} &55.8  &\underline{51.7}  &65.0  &\underline{98.3} &67.5 \\

\ding{180}&\multicolumn{1}{c}{\ding{51}}  &\ding{51}   &\ding{51}  & \multicolumn{1}{c|}{\ding{51}} &\textbf{53.8}  &\textbf{51.5}  & \textbf{64.3} &\textbf{98.1} &\textbf{66.9} \\
\bottomrule
\end{tabular}}

\end{table}

\begin{table}[htp]

\caption{
Performance comparisons with different number of annotations per batch. \ding{172} one, \ding{173} three, and \ding{174} five annotations are considered. $\mathcal{BFS}$ represents the buffer size.}
\label{tab:differen-num-oracle-lables}
\vspace{-0.1in}
\resizebox{1.0\linewidth}{!}{
\begin{tabular}{l|l|ccccc}
\toprule
    &Methods     &ImageNet-C  &ImageNet-R  &ImageNet-K  &ImageNet-A  &Avg. Err. \\  \toprule
    
\multirow{4}{*}{\ding{172}} &\multicolumn{1}{l|}{$\bullet~$Baseline} &60.6  &54.9  &66.9  &99.4 &70.5  \\
 &\multicolumn{1}{l|}{$\bullet~$SimATTA~\cite{gui2024active} ($\mathcal{BFS}=300$)} &56.6  &53.2 &65.7 &97.9 &68.4 \\
 &\multicolumn{1}{l|}{$\bullet~$Ours} &\underline{53.8}   &\underline{51.5}  &\underline{64.3}  &\underline{98.1} &\underline{66.9} \\
 &\multicolumn{1}{l|}{$\bullet~$Ours ($\mathcal{BFS}=300$)} &\textbf{52.7}  &\textbf{50.8}  &\textbf{63.3}  &\textbf{97.6} &\textbf{66.1} \\
 \midrule

 \multirow{4}{*}{\ding{173}} &\multicolumn{1}{l|}{$\bullet~$Baseline} &54.8  &52.4  &65.9  &98.4 &67.9  \\
 &\multicolumn{1}{l|}{$\bullet~$SimATTA~\cite{gui2024active} ($\mathcal{BFS}=300$)} &54.0  &51.3 &64.5 &\underline{97.2} &66.8 \\
 &\multicolumn{1}{l|}{$\bullet~$Ours} &\underline{51.9}   &\underline{49.7}  &\underline{62.3}  &\underline{97.2} &\underline{65.3} \\
 &\multicolumn{1}{l|}{$\bullet~$Ours ($\mathcal{BFS}=300$)} &\textbf{50.9}  &\textbf{48.8}  &\textbf{61.4}  &\textbf{96.1} &\textbf{64.3} \\
 \midrule
 
 \multirow{4}{*}{\ding{174}} &\multicolumn{1}{l|}{$\bullet~$Baseline} &52.9  &51.9  &65.6  &97.6 &67.0  \\
 &\multicolumn{1}{l|}{$\bullet~$SimATTA~\cite{gui2024active} ($\mathcal{BFS}=300$)} &52.2  &49.5 &62.5 &\underline{96.3} &65.1 \\
 &\multicolumn{1}{l|}{$\bullet~$Ours} &\underline{51.2}   &\underline{48.7}  &\underline{61.4}  &\underline{96.3} &\underline{64.4} \\
 &\multicolumn{1}{l|}{$\bullet~$Ours ($\mathcal{BFS}=300$)} &\textbf{50.2}  &\textbf{47.9}  &\textbf{60.7}  &\textbf{95.4} &\textbf{63.6} \\

\bottomrule
\end{tabular}}
\end{table}
The ATTA baseline consistently outperforms the non-active learning 
approach TENT~\cite{wang2020tent}, but lags behind ETA~\cite{niu2022efficient} in sequences that 
begin with `A’, `C’, and `S’.
In comparison, 
our method demonstrates significant performance improvements, achieving large gains in terms of the average accuracy by 5.7\%, 4.1\%, 4.8\%, and 3.1\% in the CTTA setting, and 3.7\%, 4.7\%, 2.9\%, and 2.0\% in the FTTA setting, compared to the ATTA baseline. Moreover, under the same annotation number and buffer settings, our method significantly outperforms SimATTA by 2.8\%, 5.1\%, 6.4\%, and 11.3\% (3.9\%, 5.4\%, 3.6\%, and 6.6\% ) on the four sequences in the CTTA (FTTA) setting, respectively.

\subsection{Ablation Studies}
In this subsection, we evaluate the effectiveness of EATTA by extensive ablation studies. All experiments adopt ResNet-50 with BatchNorm as the backbone.

\noindent\textbf{Effectiveness of Each Component}. 
In Table~\ref{tab:components}, `PD’ and `CB’ represent the prediction difference-based sample selection method and the class balancing strategy described in Section 3.2, while `GND’ and `EMA’ denote the gradient-norm based debiasing and EMA strategies introduced in Section 3.3. On ImageNet-C , applying `PD’ or `GND’ alone significantly reduces the average error rate by 1.0\% and 4.0\%, respectively. Combining `PD’ with `CB’ or `GND’ with `EMA’ further reduce the error rate by 1.3\% and 0.9\%, respectively. Moreover, combining `PD’ and `GND’ lowers the error rate on ImageNet-C by 4.4\%. Finally, the full version of EATTA achieve the best performance, achieving a 5.8\% and 3.3\% reduction in the error rate on ImageNet-C and the average error rate on four benchmarks, respectively.

\begin{table}[htp]
\caption{Performance comparisons with different number of batches to annotate one sample. \ding{172} one, \ding{173} three, and five\ding{174} batches are considered.
}
\vspace{-0.1in}
\resizebox{1.0\linewidth}{!}{
\begin{tabular}{l|l|ccccc}
\toprule
     &Methods     &ImageNet-C  &ImageNet-R  &ImageNet-K  &ImageNet-A  &Avg. Err. \\  \toprule
    
\multirow{4}{*}{\ding{172}} &\multicolumn{1}{l|}{$\bullet~$Baseline} &60.6  &54.9  &66.9  &99.4 &70.5  \\
 &\multicolumn{1}{l|}{$\bullet~$SimATTA~\cite{gui2024active} ($\mathcal{BFS}=300$)} &56.6  &53.2 &65.7  &\underline{97.9} &68.4 \\
 &\multicolumn{1}{l|}{$\bullet~$Ours} &\underline{53.8}   &\underline{51.5}  &\underline{64.3}  &98.1 &\underline{66.9} \\
 &\multicolumn{1}{l|}{$\bullet~$Ours ($\mathcal{BFS}=300$)} &\textbf{52.9}  &\textbf{50.5}  &\textbf{63.4}  &\textbf{97.6} &\textbf{66.1} \\
\midrule

 \multirow{4}{*}{\ding{173}} &\multicolumn{1}{l|}{$\bullet~$Baseline} &60.9  &57.0  &67.7  &99.7 &71.3 \\
 &\multicolumn{1}{l|}{$\bullet~$SimATTA~\cite{gui2024active} ($\mathcal{BFS}=300$)} 
 &58.9  &56.7  &68.1  &99.0 &70.7 \\
 &\multicolumn{1}{l|}{$\bullet~$Ours} &\underline{56.8}  &\textbf{53.0}  &\underline{66.0}  &\underline{98.6} &\underline{68.6} \\
 &\multicolumn{1}{l|}{$\bullet~$Ours ($\mathcal{BFS}=300$)} &\textbf{56.5}  &\underline{53.5}  &\textbf{64.7}  &\textbf{98.5} &\textbf{68.3} \\
\midrule

 \multirow{4}{*}{\ding{174}} &\multicolumn{1}{l|}{$\bullet~$Baseline} &61.2  &57.2  &68.5  &99.8 &71.7 \\
 &\multicolumn{1}{l|}{$\bullet~$SimATTA~\cite{gui2024active} ($\mathcal{BFS}=300$)} &59.5  &\underline{56.9} &68.4  &99.2 &71.0 \\
 &\multicolumn{1}{l|}{$\bullet~$Ours} &\underline{59.0}  &\textbf{53.8}  &\underline{66.2}  &\underline{98.9} & \underline{69.5}\\
 &\multicolumn{1}{l|}{$\bullet~$Ours ($\mathcal{BFS}=300$)} &\textbf{57.1}  &\textbf{53.8}  &\textbf{65.0}  &\textbf{98.5} &\textbf{68.6} \\
\bottomrule
\end{tabular}}\label{table:few-oracle}
 \vspace{-0.1in}
\end{table}
\begin{table}[htp]
\caption{Performance comparisons between different sample selection methods. All methods in this table select one sample per batch.}
\vspace{-0.1in}
\resizebox{1.0\linewidth}{!}{
\begin{tabular}{l|l|ccccc}
\toprule
     &Methods     &ImageNet-C  &ImageNet-R  &ImageNet-K  &ImageNet-A  &Avg. Err. \\  \toprule
    
\multirow{3}{*}{\rotatebox{90}{AL}} &\multicolumn{1}{l|}{$\bullet~$Max Entropy~\cite{wang2014new}} &61.6  &52.5  &66.0  &\underline{99.0} &69.8  \\
 &\multicolumn{1}{l|}{$\bullet~$Least Confidence~\cite{wang2014new}} &60.8  &52.7 &68.4  &\underline{99.0} &70.2 \\
 &\multicolumn{1}{l|}{$\bullet~$Min Margin~\cite{xie2022active}} &59.8   &\underline{51.8}  &65.1  &99.4 &69.0 \\
\midrule

 \multirow{3}{*}{\rotatebox{90}{TTA}}  &\multicolumn{1}{l|}{$\bullet~$Random} &54.7  &53.5  &65.4  &98.5 &68.0 \\
 
 &\multicolumn{1}{l|}{$\bullet~$Entropy Margin~\cite{niu2022efficient}} &54.3  &\underline{51.8}  &\underline{64.8}  &99.6 &\underline{67.6} \\
 &\multicolumn{1}{l|}{$\bullet~$PLPD~\cite{lee2024entropy}} &54.6  &55.2  &66.6  &99.5 &69.0 \\
\midrule
 \multirow{3}{*}{\rotatebox{90}{ATTA}}

 &\multicolumn{1}{l|}{$\bullet~$Incremental K-Clustering~\cite{gui2024active}} &55.4  &52.8  &67.5  &99.3 &68.8 \\
 &\multicolumn{1}{l|}{$\bullet~$Dynamic Margin~\cite{chen2024towards}} 
 &\underline{54.5}  &52.4  &65.7  &99.4 &68.0 \\
 &\multicolumn{1}{l|}{$\bullet~$Ours} &\textbf{53.8} &\textbf{51.5}  &\textbf{64.3}  &\textbf{98.1} &\textbf{66.9} \\
\bottomrule
\end{tabular}}\label{tab:different ss}
\end{table}

\noindent\textbf{Different Number of Annotations Per Batch}. We test the 
performance of our method with different number of annotations per batch, and show the results in Table~\ref{tab:differen-num-oracle-lables}. Without the help of sample buffer, our method (EATTA) outperforms SimATTA~\cite{gui2024active} with the buffer by 1.5\%, 1.5\%, and 0.7\% in terms of the average error rates for 1, 3, and 5 annotations per batch, respectively. With the help of the sample buffer, EATTA achieves performance gains by 2.3\%, 2.5\%, and 1.5\% compared with SimATTA. These results demonstrate the robustness of our method when dealing with limited number of annotations per batch.

\noindent\textbf{Different Number of Batches to Annotate One Sample}.
We further evaluate our method under more challenging conditions by reducing the number of batches to annotate one sample. The experimental results are summarized in Table~\ref{table:few-oracle}. It is shown that the performance of both the ATTA baseline and SimATTA~\cite{gui2024active}  declines significantly as the number of annotations decreases. Notably, if the annotation frequency reduces to every three or five batches, their performance becomes worse than the baseline that annotates one sample per batch. In contrast, our method remains robust, and consistently outperforms the ATTA baseline that annotates one sample per batch. Finally, with the help of the sample buffer, our approach consistently outperforms SimATTA by 2.3\%, 2.4\%, and 2.4\% in terms of the average error rate across the three scenarios.

\noindent\textbf{Comparisons between Different Sample Selection Methods}.
We compare the performance of different sample selection methods from active learning (\textit{i.e.}, AL in Table~\ref{tab:different ss}), TTA, and ATTA. To facilitate fair comparison, we replace our sample selection method with another one
\begin{table*}[htp]
    \vspace{-0.2in}
    \caption{Performance comparisons between different ATTA methods on ImageNet-C. Except for Baseline*(GT), all methods adopt the ViT-L-16 model to annotate the selected samples. The Baseline (GT) adopts ground-truth labels.}
    \vspace{-0.3in}
    \label{tab:large-model-annotate}
    \newcommand{\tabincell}[2]{\begin{tabular}{@{}#1@{}}#2\end{tabular}}
    \begin{center}
    % \subcaption{}
    \begin{threeparttable}
        \resizebox{1.0\linewidth}{!}{
            \begin{tabular}{ll|ccc|cccc|cccc|cccc|c}
               & \multicolumn{1}{c}{}  & \multicolumn{3}{c}{Noise} & \multicolumn{4}{c}{Blur} & \multicolumn{4}{c}{Weather} & \multicolumn{4}{c}{Digital}  \\
                \toprule 
            \multicolumn{1}{l|}{\textbf{C}}  &  Methods  & Gauss. & Shot & Impul. & Defoc. & Glass & Motion & Zoom & Snow & Frost & Fog & Brit. & Contr. & Elastic & Pixel & JPEG & Avg. Err.  \\
                % \cmidrule{1-18}
                \midrule
       
% \\
 \multicolumn{1}{l|}{\multirow{6}{*}{\rotatebox{90}{Active}}}&$\bullet~$Baseline*(GT) &69.3 &62.7 		 &64.4	 &72.9 	 &70.2 	 &64.7 	 &60.4 	 &63.8 	 &64.1 	 &53.3  &43.8 	 & 64.1		 &53.7	 &49.0 	 &52.3  &60.6
\\
\multicolumn{1}{l|}{}&$\bullet~$Baseline$^{*}$    &73.4  &71.1  &72.5  &78.6  &75.1  &69.4  &64.9  &66.1  &64.4  &54.7  &42.7  &65.4  &56.4  &55.4  &56.1  &64.4
\\

\multicolumn{1}{l|}{}&$\bullet~$SimATTA$^\dagger$~\cite{gui2024active}  ($\mathcal{BFS}=300$)  &73.0 &66.3 &68.5 &74.5 &70.5 &62.6 &59.4 &59.6 &58.4 &48.6 &36.2 &59.4 &51.4 &50.4 &51.9 &59.4
\\

\multicolumn{1}{l|}{} &\cellcolor{mycolor1}$\bullet~$Ours$^{*}$   &\cellcolor{mycolor1}\underline{66.3} &\cellcolor{mycolor1}59.2 	 &\cellcolor{mycolor1}60.1	 &\cellcolor{mycolor1}70.0 	 &\cellcolor{mycolor1}68.3 	 &\cellcolor{mycolor1}62.8 	 &\cellcolor{mycolor1}57.5 	 &\cellcolor{mycolor1}59.3 	 &\cellcolor{mycolor1}61.1 	 &\cellcolor{mycolor1}48.5  &\cellcolor{mycolor1}38.5 	 &\cellcolor{mycolor1}66.4 		 &\cellcolor{mycolor1}51.7 	 &\cellcolor{mycolor1}48.6 	 &\cellcolor{mycolor1}52.1  &\cellcolor{mycolor1}58.0
\\
\multicolumn{1}{l|}{}&\cellcolor{mycolor2}$\bullet~$Ours$^\dagger$   &\cellcolor{mycolor2}\textbf{65.5} &\cellcolor{mycolor2}\textbf{58.7} 		 &\cellcolor{mycolor2}\textbf{59.3}	 &\cellcolor{mycolor2}\underline{68.5} 	 &\cellcolor{mycolor2}\underline{66.3} 	 &\cellcolor{mycolor2}\underline{58.8} 	 &\cellcolor{mycolor2}\underline{54.5} 	 &\cellcolor{mycolor2}\underline{54.9} 	 &\cellcolor{mycolor2}\underline{56.5} 	 &\cellcolor{mycolor2}\underline{45.0}  &\cellcolor{mycolor2}\underline{36.3} 	 &\cellcolor{mycolor2}\underline{57.7} 		 &\cellcolor{mycolor2}\underline{46.7} 	 &\cellcolor{mycolor2}\underline{42.8} 	 &\cellcolor{mycolor2}\underline{46.1}  &\cellcolor{mycolor2}\underline{54.5}
\\
\multicolumn{1}{l|}{}&\cellcolor{mycolor3}$\bullet~$Ours$^\dagger$ ($\mathcal{BFS}=300$)   &\cellcolor{mycolor3}\underline{66.3} &\cellcolor{mycolor3}\underline{59.1} 		 &\cellcolor{mycolor3}\underline{60.0}	 &\cellcolor{mycolor3}\textbf{67.6} 	 &\cellcolor{mycolor3}\textbf{65.2}	 &\cellcolor{mycolor3}\textbf{57.4} 	 &\cellcolor{mycolor3}\textbf{53.7} 	 &\cellcolor{mycolor3}\textbf{53.8} 	 &\cellcolor{mycolor3}\textbf{55.3} 	 &\cellcolor{mycolor3}\textbf{44.1}  &\cellcolor{mycolor3}\textbf{35.4} 	 &\cellcolor{mycolor3}\textbf{55.5}		 &\cellcolor{mycolor3}\textbf{46.6} 	 &\cellcolor{mycolor3} \textbf{42.6}	 &\cellcolor{mycolor3}\textbf{46.0}  &\cellcolor{mycolor3}\textbf{53.9}
\\        
                \bottomrule
            \end{tabular}
        }
    \end{threeparttable}
    \end{center}
\vspace{-0.1in}
\end{table*}
but 
remain 
all 
the other components in EATTA. We summarize the experimental results in Table~\ref{tab:different ss}. It is shown that the uncertainty-based selection methods~\cite{wang2014new,lewis1994heterogeneous,xie2022active} are less effective for ATTA, as high-uncertainty samples are difficult for the model to learn in a single-step optimization. Similar phenomena are observed in incremental clustering~\cite{gui2024active} and PLPD-based methods~\cite{lee2024entropy}. Samples selected randomly or according to dynamic margin~\cite{chen2024towards} yield worse results than ours, probably because those selected samples are not informative enough for ATTA. In comparison, our method achieves the best performance, demonstrating the effectiveness of our effortless active labeling approach.

\subsection{Further Analysis}

\begin{figure}[tp]  
    \centering
    \includegraphics[width=1.0\linewidth]{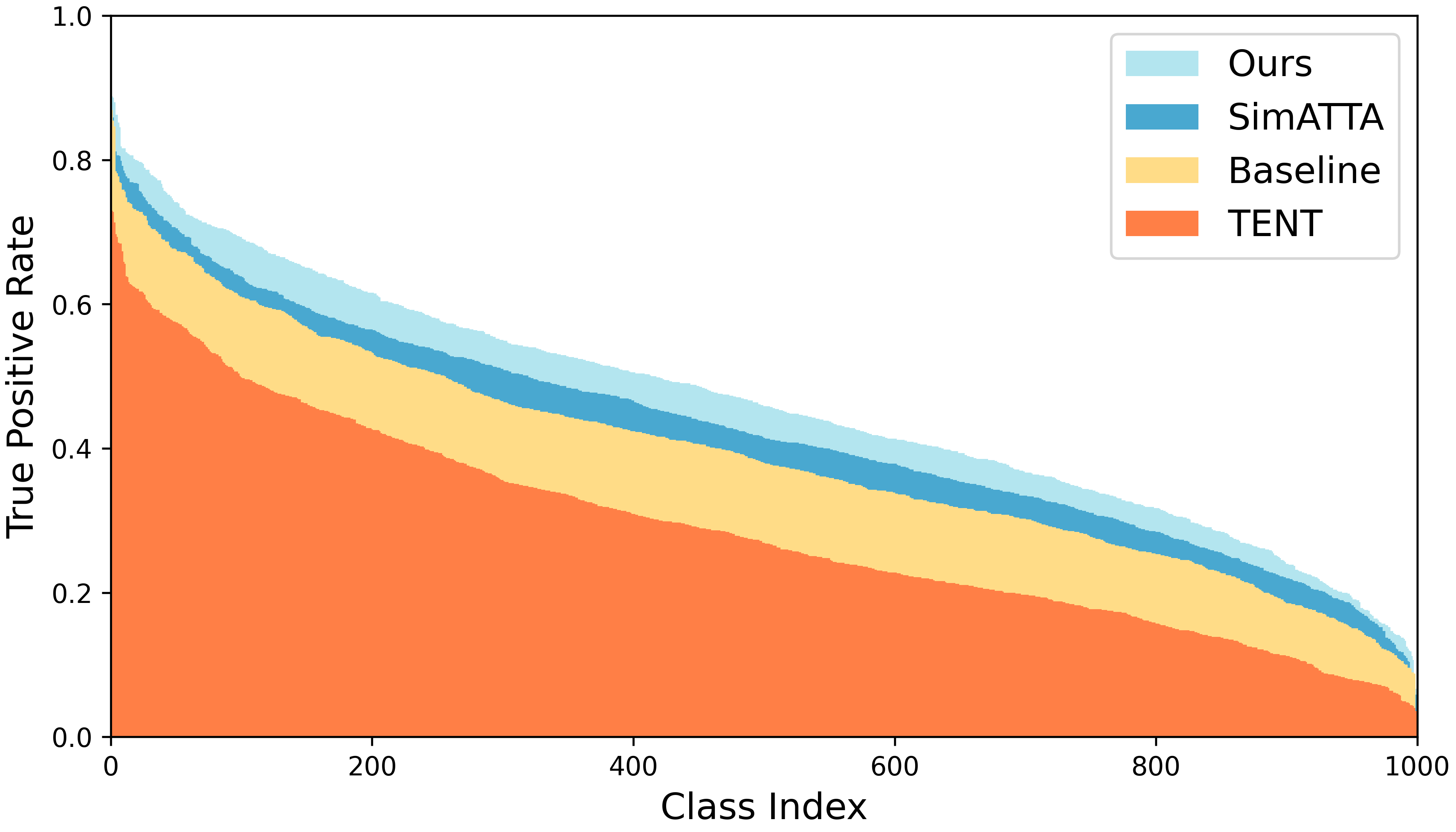}  
    \vspace{-0.25in}
    \caption{Comparisons in true positive rates on each category of the ImageNet-C database by different ATTA methods. }  
    \label{fig:cls_distribution}  
  
\end{figure}

\noindent \textbf{Comparisons in True Positive Rates between ATTA Methods}. 
In Figure~\ref{fig:cls_distribution}, we compare the true positive rate on each category of the ImageNet-C database by different ATTA methods. All ATTA methods in this figure annotate only one sample per batch. It is shown that our approach achieves the best true positive rate on all categories. This is due to two reasons. First, our approach selects samples that are optimal for the perspective of ATTA. They are both informative and easy to learn by single-step optimization. Therefore, the model can be more effectively optimized. Second, we adopt strategies to handle the class imbalance problem in the annotation process and the imbalance problem between the two training objectives. Therefore, the prediction accuracy of our approach is better than others on all the categories.

\noindent \textbf{Replacing Human Annotator with Large Models}. 
In this experiment, we explore the potential of using large models to replace human annotators. The experimental results are illustrated in Table~\ref{tab:large-model-annotate}. All ATTA methods in this table adopt the ViT-L-16 model to annotate the selected samples, except for `Baseline* (GT)’, which adopts ground-truth labels. As illustrated in Table~\ref{tab:large-model-annotate}, the performance of the baseline degrades by 3.8\% in terms of the average error rate if we conduct this replacement. While SimATTA~\cite{gui2024active} shows modest improvements over `Baseline* (GT)’, our approach reduces the average error rate by 2.6\% compared with `Baseline* (GT)’. Finally, if we annotate three samples per batch using ViT-L-16 and adopt the same sample buffer as SimATTA, our approach significantly outperforms `Baseline* (GT)’ by 6.7\% in terms of the average error rate. These experimental
\noindent results highlight EATTA’s potential to deploy in real-world applications.

\section{Conclusion and Limitations}
In this paper, we propose the EATTA approach, which relieves the increasing annotation burden in ATTA as online data scale grows by requiring only one labeled sample per batch, or even per multi-batches. 
We point out two critical problems in this context: (i) how to select the most valuable sample for labeling, and (ii) how to leverage this sample for model adaptation. 
We regard the samples that lie at the border between the source- and target-domain data distributions are the most feasible for the model to learn by one iteration, and introduce an efficient strategy to identify these samples via feature perturbation. Moreover, we introduce a gradient norm-based method to balance the training objectives between the labeled and unlabeled data. Extensive experiments on widely-used databases demonstrate that EATTA outperforms state-of-the-art methods with much less annotation cost. The same as other ATTA methods, our work involves annotations from humans or large models. Although we have significantly reduced this annotation burden, delays may still appear in the TTA process.

\noindent\textbf{Broader Impacts}. 
Test-time adaptation enables a model pretrained on source domain data to adapt to target domain data in real-time. It has broad applications in dynamic scenarios such as autonomous driving. To the best of our knowledge, our work does not have obvious negative social impacts.

% \noindent\textbf{Acknowledgement}.
\section{Acknowledgement}
This work was supported by the National Natural Science Foundation of China under Grant 62476099 and 62076101, Guangdong Basic and Applied Basic Research Foundation under Grant 2024B1515020082 and 2023A1515010007, the Guangdong Provincial Key Laboratory of Human Digital Twin under Grant 2022B1212010004, the TCL Young Scholars Program, and the 2024 Tencent AI Lab Rhino-Bird Focused Research Program.

{
    \small
    \bibliographystyle{ieeenat_fullname}
    \bibliography{main}
}
\setcounter{section}{0}
\setcounter{figure}{0}
\setcounter{table}{0}
\renewcommand{\thesection}{\Alph{section}}
\renewcommand{\thefigure}{\Alph{figure}}
\renewcommand{\thetable}{\Alph{table}}
\clearpage
\maketitlesupplementary

This supplementary material is organized into five sections: Section~\ref{sp-0} discusses the importance of border samples in TTA.
Section~\ref{sp-1} presents additional experimental results, including the substitution of human annotators with large models (Table~\ref{tab:F-human-to-model}), and the robustness of our method across various backbones (Tables~\ref{tab:imagenet-c-rn50_gn} and~\ref{tab:imagenet-c-vit}). Section~\ref{sp-2} offers further ablation studies, such as the effects of hyper-parameter variations (Table~\ref{tab:ablation-hyper-parameter}). Section~\ref{sp-3} provides detailed descriptions of the datasets used, including ImageNet-C, -A, -R, -K, and PACS. Finally, Section~\ref{sp-4} explains the pre-training protocol and implementation details of TTA and ATTA baselines.

\section{Importance of Border Samples in TTA}\label{sp-0}
TTA adapts a pre-trained model in real-time based on online target data with significant domain shifts, which presents challenges from both noisy labels and efficiency requirements. ATTA mitigates the adverse effect of noisy labels by introducing human/large model annotations of a small set of samples. However, as we demonstrate in Figure~\ref{fig:ent_diff}, these samples could be difficult for the model to learn, thereby hindering adequate adaptation.
To address this problem, we propose to select samples that are both informative and feasible to learn from a single-step optimization perspective. 
Specifically, we prefer samples that are distributed at the border between the source and target domains, as they provide learnable target domain knowledge and help the model gradually adapt to the target domain~\cite{kumar2020understanding,chen2021gradual}.

\begin{figure}[hp] 
    \centering
    \includegraphics[width=\linewidth]{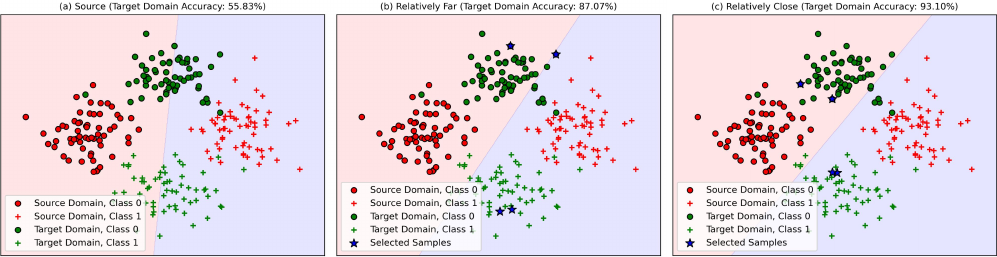}
     \caption{Performance comparisons with two sampling strategies for model adaptation on a toy example. (a) The initial decision boundary of the source model. (b) The updated decision boundary using data points that are relatively far from the source domain. (c) The updated decision boundary using data points that are relatively close to the source domain.}  
    \label{fig:toy_example}  
\end{figure}

Furthermore, we provide a toy example to compare two sampling strategies, where the source model is fine-tuned with data points (blue stars) that are far from and close to the source domain, respectively.
To this end, we first train a toy model on red data and test it on green data. We then select two data points per class in the green data to fine-tune the pre-trained toy model and visualize the decision boundary in Figure~\ref{fig:toy_example}. 

As shown in Figure~\ref{fig:toy_example}, the model performs better when fine-tuned on data points close to the source domain, achieving 93.10\%, compared to 87.07\% when fine-tuned on points far from the source domain. This demonstrate border samples are the optimal ones to facilitate robust adaptation.

\section{More Experimental Results}\label{sp-1}

\textbf{Replacing Human Annotator with Large Models}. This experiment is an extension of Table 8 of the main paper. We perform experiments in the fully test-time adaptation setting and present the results in Table~\ref{tab:F-human-to-model}. SimATTA~\cite{gui2024active} shows modest improvements over the baseline where a large model provides the annotations. However, it underperforms compared to the baseline where human experts provide annotations. In contrast, our method consistently outperforms the baseline, regardless of whether the annotations come from human experts or large models. This highlights the potential of our method to deploy in real-world applications.

\noindent\textbf{Performance across Various Backbones}. We provide experimental results on ImageNet-C for continual and fully test-time adaptation settings using ResNet-50 with GroupNorm (Table~\ref{tab:imagenet-c-rn50_gn}) and ViT-B-16 (Table~\ref{tab:imagenet-c-vit}), respectively. 
As shown in Table~\ref{tab:imagenet-c-rn50_gn}, the basic version of our method (\textit{i.e.}, Ours*) surpasses the baseline by 1.4\% and 1.1\% for CTTA and FTTA, respectively. Moreover, under the same annotation and buffer setting, our method outperforms SimATTA~\cite{gui2024active} by 6.3\% and 4.4\% for CTTA and FTTA, respectively.
As shown in Table~\ref{tab:imagenet-c-vit},  the basic version of our method surpasses the baseline by 1.9\% and 0.4\% for CTTA and FTTA, respectively. Moreover, under the same annotation and buffer setting, our method outperforms SimATTA by 3.1\% and 3.6\% for CTTA and FTTA, respectively. This demonstrates the robustness of our method across various backbones.

\begin{table*}[tp]
    \vspace{-0.1in}
    \caption{Performance comparisons between different ATTA methods on ImageNet-C for fully test-time adaptation setting (\textit{i.e.}, `\textbf{F}'). Except for Baseline*(GT), all methods adopt the ViT-L-16 model to annotate the selected samples. The Baseline*(GT) adopts ground-truth labels. ``*'' and ``$\dagger$'' indicate 1 and 3 samples are annotated per batch, respectively. $\mathcal{BFS}$ is the buffer size. The \textbf{best} and \underline{second-best} performances are highlighted.}
    \vspace{-0.2in}
    \label{tab:F-human-to-model}
    \newcommand{\tabincell}[2]{\begin{tabular}{@{}#1@{}}#2\end{tabular}}
    \begin{center}
    % \subcaption{}
    \begin{threeparttable}
        \resizebox{1.0\linewidth}{!}{
            \begin{tabular}{ll|ccc|cccc|cccc|cccc|c}
               & \multicolumn{1}{c}{}  & \multicolumn{3}{c}{Noise} & \multicolumn{4}{c}{Blur} & \multicolumn{4}{c}{Weather} & \multicolumn{4}{c}{Digital}  \\
                \toprule 
            \multicolumn{1}{l|}{\textbf{F}}  &  Methods  & Gauss. & Shot & Impul. & Defoc. & Glass & Motion & Zoom & Snow & Frost & Fog & Brit. & Contr. & Elastic & Pixel & JPEG & Avg. Err.  \\
                % \cmidrule{1-18}
                \midrule
       
% \\
 \multicolumn{1}{l|}{\multirow{6}{*}{\rotatebox{90}{Active}}}&$\bullet~$Baseline$^{*}$(GT) &69.3 &67.5 		 &68.1	 &72.6 	 &70.2 	 &57.7 	 &51.2 	 &52.3 	 &57.7 	 &43.0  &33.3 	 &63.0 		 &45.6 	 &42.1 	 &47.7  &56.1
\\
\multicolumn{1}{l|}{}&$\bullet~$Baseline$^{*}$    &73.8  &71.4  &71.6  &74.3  &75.3  &64.1  &59.0  &57.1  &59.0  &46.4  &33.8  &65.6  &51.4  &54.0  &51.4  &60.6
\\
\multicolumn{1}{l|}{}&$\bullet~$SimATTA$^\dagger$~\cite{gui2024active}  ($\mathcal{BFS}=300$)  &71.1 &68.8 &69.9 &72.7 &75.6 &62.4 &59.5 &56.1 &57.4 &46.6 &33.2 &81.7 &52.5 &47.4 &49.3 &60.3
\\

\multicolumn{1}{l|}{} &\cellcolor{mycolor1}$\bullet~$Ours$^{*}$   &\cellcolor{mycolor1}67.3 &\cellcolor{mycolor1}65.4 	 &\cellcolor{mycolor1}\underline{66.1}	 &\cellcolor{mycolor1}69.5 	 &\cellcolor{mycolor1}70.1 	 &\cellcolor{mycolor1}\underline{56.8} 	 &\cellcolor{mycolor1}\textbf{50.6} 	 &\cellcolor{mycolor1}\underline{50.8} 	 &\cellcolor{mycolor1}\underline{56.0} 	 &\cellcolor{mycolor1}\underline{42.4}  &\cellcolor{mycolor1}\underline{32.7} 	 &\cellcolor{mycolor1}63.7 		 &\cellcolor{mycolor1}\textbf{44.3} 	 &\cellcolor{mycolor1}\textbf{40.9} 	 &\cellcolor{mycolor1}46.2  &\cellcolor{mycolor1}54.9
\\
\multicolumn{1}{l|}{}&\cellcolor{mycolor2}$\bullet~$Ours$^\dagger$   &\cellcolor{mycolor2}\textbf{66.7} &\cellcolor{mycolor2}\underline{65.1}		 &\cellcolor{mycolor2}66.2	 &\cellcolor{mycolor2}\underline{69.0} 	 &\cellcolor{mycolor2}\underline{69.5}	 &\cellcolor{mycolor2}\underline{56.8}	 &\cellcolor{mycolor2}\underline{50.8} 	 &\cellcolor{mycolor2}\textbf{50.7}	 &\cellcolor{mycolor2}\underline{56.0} 	 &\cellcolor{mycolor2}\underline{42.4}  &\cellcolor{mycolor2}\underline{32.7}	 &\cellcolor{mycolor2}\underline{61.1} 		 &\cellcolor{mycolor2}\textbf{44.5} 	 &\cellcolor{mycolor2}\textbf{41.3} 	 &\cellcolor{mycolor2}\textbf{46.2}  &\cellcolor{mycolor2}\underline{54.6}
\\
\multicolumn{1}{l|}{}&\cellcolor{mycolor3}$\bullet~$Ours$^\dagger$ ($\mathcal{BFS}=300$)   &\cellcolor{mycolor3}\underline{66.9}
&\cellcolor{mycolor3}\textbf{64.9}		 &\cellcolor{mycolor3}\textbf{65.8}	 &\cellcolor{mycolor3}\textbf{68.9} 	 &\cellcolor{mycolor3}\textbf{69.0}	 &\cellcolor{mycolor3}\textbf{56.1}	 &\cellcolor{mycolor3}51.6 	 &\cellcolor{mycolor3}\textbf{50.7} 	 &\cellcolor{mycolor3}\textbf{55.3}	 &\cellcolor{mycolor3}\textbf{42.1}  &\cellcolor{mycolor3}\textbf{32.4} 	 &\cellcolor{mycolor3}\textbf{59.4}		 &\cellcolor{mycolor3}44.6 	 &\cellcolor{mycolor3}41.8	 &\cellcolor{mycolor3}\underline{46.3} &\cellcolor{mycolor3}\textbf{54.4}
\\        
                \bottomrule
            \end{tabular}
        }
    \end{threeparttable}
    \end{center}

\end{table*}

\begin{table*}[htp]
    \vspace{-0.1in}
    \caption{Performance comparisons on ImageNet-C for continual (\textit{i.e.}, \textbf{C}) and fully (\textit{i.e.}, \textbf{F}) test-time adaptation settings. The backbone is ResNet-50 with GroupNorm. $\mathcal{BFS}$ is the buffer size.
    The \textbf{best} and \underline{second-best} performances are highlighted.}
    \vspace{-0.2in}
    \label{tab:imagenet-c-rn50_gn}
    \newcommand{\tabincell}[2]{\begin{tabular}{@{}#1@{}}#2\end{tabular}}
    \begin{center}
    \begin{threeparttable}
        \resizebox{1.0\linewidth}{!}{
            \begin{tabular}{ll|ccc|cccc|cccc|cccc|c}
               & \multicolumn{1}{c}{}  & \multicolumn{3}{c}{Noise} & \multicolumn{4}{c}{Blur} & \multicolumn{4}{c}{Weather} & \multicolumn{4}{c}{Digital}  \\
                \toprule 
            \multicolumn{1}{l|}{\textbf{C}}  &  Methods  & Gauss. & Shot & Impul. & Defoc. & Glass & Motion & Zoom & Snow & Frost & Fog & Brit. & Contr. & Elastic & Pixel & JPEG & Avg. Err.  \\
                \midrule
        \multicolumn{1}{l|}{\multirow{5}{*}{\rotatebox{90}{Non-Active}}}      &  $\bullet$ Source  &82.0 &80.2 		 &82.1	 &80.3 	 &88.7 	 &78.6 	 &75.1 	 &59.6 	 &52.7 	 &66.4  &30.7 	 &63.7 		 &81.4 	 &71.6 	 &47.7  &69.4  \\
\multicolumn{1}{l|}{}& $\bullet~$TENT~\cite{wang2020tent}   &95.1 &99.6 		 &99.8	 &95.6 	 &99.8 	 &99.7 	 &99.7 	 &99.2 	 &99.8 	 &99.8  &99.0 	 &99.9 		 &99.9 	 &99.9 	 &99.4  &99.1
\\
\multicolumn{1}{l|}{}& $\bullet~$CoTTA~\cite{wang2022continual}   &89.8 &73.8 		 &82.1	 &87.9 	 &82.9 	 &80.8 	 &76.3 	 &82.1 	 &74.5 	 &73.4  &55.0 	 &75.6 		 &78.5 	 &56.1 	 &60.8  &76.0
\\
\multicolumn{1}{l|}{}&$\bullet~$SAR~\cite{niu2023towards}   &71.8 &58.2 		 &56.1	 &83.6 	 &80.0 	 &86.3 	 &96.8 	 &98.9 	 &70.4 	 &50.0  &29.6 	 &53.4 		 &86.2 	 &93.7 	 &98.1  &74.2
\\
\multicolumn{1}{l|}{}&$\bullet~$ETA~\cite{niu2022efficient}   &64.1 &59.3		 &60.9	 &77.8 	 &73.7 	 &71.5 	 &62.7 	 &60.5 	 &56.3 	 &52.4  &38.1 	 &53.5 		 &60.2 	 &54.4 	 &47.1  &59.5
\\
\midrule
 \multicolumn{1}{l|}{\multirow{5}{*}{\rotatebox{90}{Active}}}&$\bullet~$Baseline$^{*}$  &62.4 &53.5 		 &53.9	 &71.6 	 &64.6 	 &59.4 	 &54.4 	 &51.5 	 &45.4 	 &41.8  &29.9 	 &46.7		 &53.2	 &43.0 	 &39.1  &51.3
\\
\multicolumn{1}{l|}{}&$\bullet~$SimATTA$^\dagger$~\cite{gui2024active}  ($\mathcal{BFS}=300$)  &63.0 &53.1		 &52.8	 &75.1 	 &69.7 	 &64.0	 &56.9 	 &54.9	 &45.6	 &45.6  &31.2 	 &48.8 		 &61.1 	 &45.8 	 &39.1  &53.8
\\

\multicolumn{1}{l|}{} &\cellcolor{mycolor1}$\bullet~$Ours$^*$   &\cellcolor{mycolor1}61.7 &\cellcolor{mycolor1}51.9 	 &\cellcolor{mycolor1}52.4	 &\cellcolor{mycolor1}71.8 	 &\cellcolor{mycolor1}63.6 	 &\cellcolor{mycolor1}58.3 	 &\cellcolor{mycolor1}52.2 	 &\cellcolor{mycolor1}48.8 	 &\cellcolor{mycolor1}43.8 	 &\cellcolor{mycolor1}40.2  &\cellcolor{mycolor1}28.3	 &\cellcolor{mycolor1}45.1 		 &\cellcolor{mycolor1}50.8 	 &\cellcolor{mycolor1}40.8 	 &\cellcolor{mycolor1}37.9  &\cellcolor{mycolor1}49.9
\\
\multicolumn{1}{l|}{}&\cellcolor{mycolor2}$\bullet~$Ours$^\dagger$   &\cellcolor{mycolor2}\underline{61.1} &\cellcolor{mycolor2}\underline{50.7} 		 &\cellcolor{mycolor2}\underline{50.6}	 &\cellcolor{mycolor2}\underline{69.8}	 &\cellcolor{mycolor2}\underline{63.1}	 &\cellcolor{mycolor2}\underline{56.6} 	 &\cellcolor{mycolor2}\underline{51.1} 	 &\cellcolor{mycolor2}\underline{47.9}	 &\cellcolor{mycolor2}\underline{43.1} 	 &\cellcolor{mycolor2}\underline{39.4}  &\cellcolor{mycolor2}\underline{27.8}	 &\cellcolor{mycolor2}\textbf{44.2} 		 &\cellcolor{mycolor2}\underline{50.0} 	 &\cellcolor{mycolor2}\underline{40.6}	 &\cellcolor{mycolor2}\underline{37.0}  &\cellcolor{mycolor2}\underline{48.9}
\\
\multicolumn{1}{l|}{}&\cellcolor{mycolor3}$\bullet~$Ours$^\dagger$ ($\mathcal{BFS}=300$)   &\cellcolor{mycolor3}\textbf{58.6} &\cellcolor{mycolor3}\textbf{49.5} 		 &\cellcolor{mycolor3}\textbf{49.7}	 &\cellcolor{mycolor3}\textbf{67.8} 	 &\cellcolor{mycolor3}\textbf{60.5}	 &\cellcolor{mycolor3}\textbf{54.9} 	 &\cellcolor{mycolor3}\textbf{49.0} 	 &\cellcolor{mycolor3}\textbf{46.5} 	 &\cellcolor{mycolor3}\textbf{42.2} 	 &\cellcolor{mycolor3}\textbf{38.2}  &\cellcolor{mycolor3}\textbf{27.5} 	 &\cellcolor{mycolor3}\underline{44.3}		 &\cellcolor{mycolor3}\textbf{47.6} 	 &\cellcolor{mycolor3}\textbf{39.3}	 &\cellcolor{mycolor3}\textbf{36.6} &\cellcolor{mycolor3}\textbf{47.5}
\\        
                \bottomrule
            \end{tabular}
        }
    \end{threeparttable}
    \end{center}

    \begin{center}
    \vspace{-0.1in}
    \begin{threeparttable}
        \resizebox{1.0\linewidth}{!}{
            \begin{tabular}{ll|ccc|cccc|cccc|cccc|c}
               & \multicolumn{1}{c}{}  & \multicolumn{3}{c}{Noise} & \multicolumn{4}{c}{Blur} & \multicolumn{4}{c}{Weather} & \multicolumn{4}{c}{Digital}  \\
                \toprule 
          \multicolumn{1}{l|}{\textbf{F}}    &  Methods  & Gauss. & Shot & Impul. & Defoc. & Glass & Motion & Zoom & Snow & Frost & Fog & Brit. & Contr. & Elastic & Pixel & JPEG & Avg. Err.  \\
                % \cmidrule{1-18}
                \midrule
              \multicolumn{1}{l|}{\multirow{5}{*}{\rotatebox{90}{Non-Active}}}   &$\bullet$ Source  &82.0 &80.2 		 &82.1	 &80.3 	 &88.7 	 &78.6 	 &75.1 	 &59.6 	 &52.7 	 &66.4  &30.7 	 &63.7 		 &81.4 	 &71.6 	 &47.7  &69.4  \\
\multicolumn{1}{l|}{}&$\bullet~$TENT~\cite{wang2020tent}   &95.1 &93.9 		 &94.2	 &85.2 	 &89.7 	 &77.7 	 &77.8 	 &73.0 	 &65.7 	 &96.8  &29.7 	 &57.8		 &88.6 	 &51.8 	 &45.6  &74.8
\\

\multicolumn{1}{l|}{}&$\bullet~$CoTTA~\cite{wang2022continual}   &97.5 &69.0		 &68.6	 &84.6 	 &83.7 	 &79.2 	 &72.7	 &70.9 	 &52.9 	 &96.1  &32.5 	 &77.9 		 &83.9 	 &55.6 	 &47.2  &71.5
\\
\multicolumn{1}{l|}{}&$\bullet~$SAR~\cite{niu2023towards}   &71.7 &68.8		 &70.1	 &81.4 	 &81.3 	 &69.4 	 &69.7 	 &59.0 	 &56.8 	 &95.1  &29.3	 &56.3 		 &82.7 	 &51.3 	 &44.8  &65.8
\\
\multicolumn{1}{l|}{}& $\bullet~$ETA~\cite{niu2022efficient}  &63.7 &61.7		 &62.9	 &72.1 	 &71.7	 &63.9 	 &61.2 	 &52.1	 &51.9 	 &45.8  &29.4	 &52.6		 &58.6 	 &45.0 	 &43.9 &55.8
\\
\midrule
\multicolumn{1}{l|}{\multirow{5}{*}{\rotatebox{90}{Active}}} &$\bullet~$Baseline$^{*}$  &62.3 &60.7		 &61.5	 &69.0 	 &71.0 	 &62.1 	 &58.5 	 &48.3 	 &47.2 	 &45.0  &27.8 	 &49.7 		 &56.3 	 &43.9 	 &42.0  &53.7
\\
\multicolumn{1}{l|}{}&$\bullet~$SimATTA$^\dagger$($\mathcal{BFS}=300$)~\cite{gui2024active}  &62.2 &60.7 		 &61.4	 &70.5 	 &71.7 	 &63.2 	 &59.2 	 &48.1	 &46.1 	 &45.0  &27.9 	 &51.4 		 &59.0	 &47.2 	 &42.1  &54.4
\\
\multicolumn{1}{l|}{}&\cellcolor{mycolor1}$\bullet~$Ours$^{*}$   &\cellcolor{mycolor1}62.2  &\cellcolor{mycolor1}59.3 	 &\cellcolor{mycolor1}61.1	 &\cellcolor{mycolor1}67.8 	 &\cellcolor{mycolor1}69.6 	 &\cellcolor{mycolor1}60.7 	 &\cellcolor{mycolor1}57.7 	 &\cellcolor{mycolor1}47.1 	 &\cellcolor{mycolor1}46.9 	 &\cellcolor{mycolor1}42.1  &\cellcolor{mycolor1}\underline{27.1} 	 &\cellcolor{mycolor1}48.7 		 &\cellcolor{mycolor1}54.4 	 &\cellcolor{mycolor1}43.1 	 &\cellcolor{mycolor1}41.0  &\cellcolor{mycolor1}52.6
\\
\multicolumn{1}{l|}{}&\cellcolor{mycolor2}$\bullet~$Ours$^\dagger$   &\cellcolor{mycolor2}\underline{61.1}  &\cellcolor{mycolor2}\underline{58.9} 		 &\cellcolor{mycolor2}\underline{60.0}	 &\cellcolor{mycolor2}\underline{67.3}	 &\cellcolor{mycolor2}\underline{69.1} 	 &\cellcolor{mycolor2}\underline{60.1} 	 &\cellcolor{mycolor2}\underline{57.3} 	 &\cellcolor{mycolor2}\underline{46.3} 	 &\cellcolor{mycolor2}\underline{46.5} 	 &\cellcolor{mycolor2}\underline{42.0}  &\cellcolor{mycolor2}\underline{27.1}	 &\cellcolor{mycolor2}\underline{48.4}		 &\cellcolor{mycolor2}\underline{54.0} 	 &\cellcolor{mycolor2}\underline{42.7} 	 &\cellcolor{mycolor2}\underline{41.0}  &\cellcolor{mycolor2}\underline{52.1}
\\
\multicolumn{1}{l|}{}&\cellcolor{mycolor3}$\bullet~$Ours$^\dagger$($\mathcal{BFS}=300$)   &\cellcolor{mycolor3}\textbf{58.8}  &\cellcolor{mycolor3}\textbf{56.3} 		 &\cellcolor{mycolor3}\textbf{57.2}	 &\cellcolor{mycolor3}\textbf{66.3}	 &\cellcolor{mycolor3}\textbf{65.7}	 &\cellcolor{mycolor3}\textbf{58.1} 	 &\cellcolor{mycolor3}\textbf{54.4} 	 &\cellcolor{mycolor3}\textbf{44.8} 	 &\cellcolor{mycolor3}\textbf{44.7} 	 &\cellcolor{mycolor3}\textbf{40.7}  &\cellcolor{mycolor3}\textbf{26.4} 	 &\cellcolor{mycolor3}\textbf{46.7}		 &\cellcolor{mycolor3}\textbf{50.6} 	 &\cellcolor{mycolor3}\textbf{40.8} 	 &\cellcolor{mycolor3}\textbf{39.3}  &\cellcolor{mycolor3}\textbf{50.0}
\\        
                \bottomrule
            \end{tabular}
        }
    \end{threeparttable}
    \end{center}
\end{table*}

\begin{table*}[htp]
    \vspace{-0.1in}
    \caption{Performance comparisons on ImageNet-C for continual (\textit{i.e.}, \textbf{C}) and fully (\textit{i.e.}, \textbf{F}) test-time adaptation settings. The backbone is ViT-B-16. $\mathcal{BFS}$ is the buffer size.
    The \textbf{best} and \underline{second-best} performances are highlighted.}
    \vspace{-0.2in}
    \label{tab:imagenet-c-vit}
    \newcommand{\tabincell}[2]{\begin{tabular}{@{}#1@{}}#2\end{tabular}}
    \begin{center}
    \begin{threeparttable}
        \resizebox{1.0\linewidth}{!}{
            \begin{tabular}{ll|ccc|cccc|cccc|cccc|c}
               & \multicolumn{1}{c}{}  & \multicolumn{3}{c}{Noise} & \multicolumn{4}{c}{Blur} & \multicolumn{4}{c}{Weather} & \multicolumn{4}{c}{Digital}  \\
                \toprule 
            \multicolumn{1}{l|}{\textbf{C}}  &  Methods  & Gauss. & Shot & Impul. & Defoc. & Glass & Motion & Zoom & Snow & Frost & Fog & Brit. & Contr. & Elastic & Pixel & JPEG & Avg. Err.  \\
                \midrule
        \multicolumn{1}{l|}{\multirow{5}{*}{\rotatebox{90}{Non-Active}}}      &  $\bullet$ Source  &66.0 &66.8 &65.0	 &68.5 	 &74.7 	 &64.0 	 &66.9 	 &57.3 	 &45.0 	 &49.4  &28.7 	 &81.8 		 &57.8 	 &60.8 	 &49.9  &60.2  \\
\multicolumn{1}{l|}{}& $\bullet~$TENT~\cite{wang2020tent}   &58.6 &54.0 		 &57.8	 &59.7 	 &77.4 	 &99.7 	 &99.8 	 &99.7 	 &99.9 	 &99.9  &99.8 	 &99.9 		 &99.9 	 &99.9 	 &99.9  &87.1
\\
\multicolumn{1}{l|}{}& $\bullet~$CoTTA~\cite{wang2022continual}   &63.5 &64.2 		 &70.4	 &93.8 	 &84.5 	 &85.8 	 &78.6 	 &99.6 	 &99.9 	 &99.9  &99.8 	 &99.9 		 &99.9 	 &99.9 	 &99.9  &89.3
\\
\multicolumn{1}{l|}{}&$\bullet~$SAR~\cite{niu2023towards}   &54.9 &50.3 		 &52.2	 &57.7 	 &60.3 	 &60.9 	 &49.6 	 &60.0 	 &48.1 	 &35.0  &26.4 	 &43.5 		 &51.7 	 &44.5 	 &35.9  &48.7
\\
\multicolumn{1}{l|}{}&$\bullet~$ETA~\cite{niu2022efficient}   &54.2 &50.3		 &52.3	 &57.5 	 &53.7	 &50.2 	 &50.8 	 &50.7	 &47.2 	 &44.4  &31.4	 &57.8		 &44.6 	 &42.4 	 &41.7 &48.6
\\
\midrule
 \multicolumn{1}{l|}{\multirow{5}{*}{\rotatebox{90}{Active}}}&$\bullet~$Baseline$^{*}$  &55.0 &49.4		 &49.9	 &54.7 	 &50.3 	 &46.4 	 &46.1 	 &46.9 	 &41.0 	 &39.4  &28.2 	 &48.6 		 &41.1 	 &38.3 	 &37.5  &44.9
\\
\multicolumn{1}{l|}{}&$\bullet~$SimATTA$^\dagger$($\mathcal{BFS}=300$)~\cite{gui2024active}  &55.2 &49.0		 &48.9	 &54.2 	 &50.6 	 &45.4	 &46.0 	 &46.4	 &39.1	 &36.9  &26.7 	 &43.7 		 &42.0 	 &38.3 	 &36.6  &43.9
\\

\multicolumn{1}{l|}{} &\cellcolor{mycolor1}$\bullet~$Ours$^{*}$   &\cellcolor{mycolor1}54.5 &\cellcolor{mycolor1}48.7 	 &\cellcolor{mycolor1}48.7	 &\cellcolor{mycolor1}53.9 	 &\cellcolor{mycolor1}49.5 	 &\cellcolor{mycolor1}44.1 	 &\cellcolor{mycolor1}43.4 	 &\cellcolor{mycolor1}46.3 	 &\cellcolor{mycolor1}38.8 	 &\cellcolor{mycolor1}36.1  &\cellcolor{mycolor1}\underline{26.4} 	 &\cellcolor{mycolor1}46.0 		 &\cellcolor{mycolor1}38.2 	 &\cellcolor{mycolor1}34.7 	 &\cellcolor{mycolor1}35.2  &\cellcolor{mycolor1}43.0
\\

\multicolumn{1}{l|}{}&\cellcolor{mycolor2}$\bullet~$Ours$^\dagger$   &\cellcolor{mycolor2}\underline{53.9} &\cellcolor{mycolor2}\underline{47.5} 		 &\cellcolor{mycolor2}\underline{47.4}	 &\cellcolor{mycolor2}\underline{52.1} 	 &\cellcolor{mycolor2}\underline{47.8} 	 &\cellcolor{mycolor2}\underline{42.6} 	 &\cellcolor{mycolor2}\underline{42.2} 	 &\cellcolor{mycolor2}\underline{44.0} 	 &\cellcolor{mycolor2}\underline{38.0}	 &\cellcolor{mycolor2}\underline{34.6}  &\cellcolor{mycolor2}\textbf{25.6} 	 &\cellcolor{mycolor2}\underline{41.7} 		 &\cellcolor{mycolor2}\underline{37.1} 	 &\cellcolor{mycolor2}\underline{34.1} 	 &\cellcolor{mycolor2}\underline{34.5}  &\cellcolor{mycolor2}\underline{41.5}
\\
\multicolumn{1}{l|}{}&\cellcolor{mycolor3}$\bullet~$Ours$^\dagger$($\mathcal{BFS}=300$)   &\cellcolor{mycolor3}\textbf{53.0} &\cellcolor{mycolor3}\textbf{46.7} 		 &\cellcolor{mycolor3}\textbf{46.9}	 &\cellcolor{mycolor3}\textbf{50.9} 	 &\cellcolor{mycolor3}\textbf{46.8} &\cellcolor{mycolor3}\textbf{42.0}	 &\cellcolor{mycolor3}\textbf{40.6} 	 &\cellcolor{mycolor3}\textbf{43.0} 	 &\cellcolor{mycolor3}\textbf{37.7} 	 &\cellcolor{mycolor3}\textbf{33.7}  &\cellcolor{mycolor3}\textbf{25.6} 	 &\cellcolor{mycolor3}\textbf{41.4}	 &\cellcolor{mycolor3}\textbf{36.0}	 &\cellcolor{mycolor3}\textbf{33.6}	 &\cellcolor{mycolor3}\textbf{33.6}  &\cellcolor{mycolor3}\textbf{40.8}
\\        
                \bottomrule
            \end{tabular}
        }
    \end{threeparttable}
    \end{center}

    \begin{center}
    \vspace{-0.1in}
    \begin{threeparttable}
        \resizebox{1.0\linewidth}{!}{
            \begin{tabular}{ll|ccc|cccc|cccc|cccc|c}
               & \multicolumn{1}{c}{}  & \multicolumn{3}{c}{Noise} & \multicolumn{4}{c}{Blur} & \multicolumn{4}{c}{Weather} & \multicolumn{4}{c}{Digital}  \\
                \toprule 
          \multicolumn{1}{l|}{\textbf{F}}    &  Methods  & Gauss. & Shot & Impul. & Defoc. & Glass & Motion & Zoom & Snow & Frost & Fog & Brit. & Contr. & Elastic & Pixel & JPEG & Avg. Err.  \\
                \midrule
              \multicolumn{1}{l|}{\multirow{5}{*}{\rotatebox{90}{Non-Active}}}   &$\bullet$ Source  &66.0 &66.8 &65.0	 &68.5 	 &74.7 	 &64.0 	 &66.9 	 &57.3 	 &45.0 	 &49.4  &28.7 	 &81.8 		 &57.8 	 &60.8 	 &49.9  &60.2  \\
\multicolumn{1}{l|}{}&$\bullet~$TENT~\cite{wang2020tent}   &58.6 &56.3 		 &56.8	 &58.1 	 &61.7 	 &52.3 	 &56.6 	 &72.2 	 &43.5 	 &93.2  &26.7 	 &54.5		 &50.0 	 &41.8 	 &41.7  &54.9
\\
\multicolumn{1}{l|}{}&$\bullet~$CoTTA~\cite{wang2022continual}   &63.0 &78.6		 &71.4	 &96.3 	 &95.3 	 &97.8 	 &88.5	 &97.9 	 &94.8 	 &45.9  &94.0 	 &98.0 		 &78.1 	 &94.6 	 &50.8  &83.0

\\
\multicolumn{1}{l|}{}&$\bullet~$SAR~\cite{niu2023towards}   &54.9 &52.9		 &53.8	 &53.4 	 &53.9 	 &46.0 	 &50.4 	 &54.3 	 &40.5 	 &38.3  &25.7 	 &55.0 		 &41.4 	 &35.9 	 &37.0  &46.2
\\
\multicolumn{1}{l|}{}& $\bullet~$ETA~\cite{niu2022efficient}  &54.2 &52.4		 &53.3	 &52.6 	 &51.6 	 &45.1 	 &46.1 	 &41.9 	 &40.1 	 &36.0  &25.5 	 &46.5 		 &38.7 	 &35.0 	 &36.2  &43.7
\\
\midrule
\multicolumn{1}{l|}{\multirow{5}{*}{\rotatebox{90}{Active}}} &$\bullet~$Baseline$^{*}$   &55.0  &53.0 	 &53.8	 &53.1 	 &53.1 	 &46.8 	 &48.6 	 &43.7 
&39.3 	&35.6
&26.0  &46.5 	 &40.9 		 &36.8 	 &38.1 	 &44.7 
\\
\multicolumn{1}{l|}{}&$\bullet~$SimATTA$^\dagger$($\mathcal{BFS}=300$)~\cite{gui2024active}  &55.2 &53.2 		 &54.0	 &55.1 	 &54.8 	 &49.1 	 &50.9 	 &44.9	 &39.6 &36.5	 &26.5  &48.1 	 &44.0 		 &39.6	 &39.1 	   &46.0
\\
\multicolumn{1}{l|}{}&\cellcolor{mycolor1}$\bullet~$Ours$^{*}$   &\cellcolor{mycolor1}54.5 &\cellcolor{mycolor1}52.6 		 &\cellcolor{mycolor1}53.6	 &\cellcolor{mycolor1}53.4 	 &\cellcolor{mycolor1}53.2 	 &\cellcolor{mycolor1}46.2 	 &\cellcolor{mycolor1}47.8 	 &\cellcolor{mycolor1}43.5 	 &\cellcolor{mycolor1}39.4 	 &\cellcolor{mycolor1}36.5  &\cellcolor{mycolor1}25.8 	 &\cellcolor{mycolor1}46.4		 &\cellcolor{mycolor1}39.5 	 &\cellcolor{mycolor1}35.8 &\cellcolor{mycolor1}36.9  &\cellcolor{mycolor1}44.3
\\
\multicolumn{1}{l|}{}&\cellcolor{mycolor2}$\bullet~$Ours$^\dagger$    &\cellcolor{mycolor2}\underline{53.9} &\cellcolor{mycolor2}\underline{52.0} 		 &\cellcolor{mycolor2}\underline{52.9}	 &\cellcolor{mycolor2}\underline{52.1} 	 &\cellcolor{mycolor2}\underline{52.4} 	 &\cellcolor{mycolor2}\underline{45.4} 	 &\cellcolor{mycolor2}\underline{47.3} 	 &\cellcolor{mycolor2}\underline{42.3} 	 &\cellcolor{mycolor2}\underline{38.7} 	 &\cellcolor{mycolor2}\underline{35.8}  &\cellcolor{mycolor2}\underline{25.4} 	 &\cellcolor{mycolor2}\underline{44.1}		 &\cellcolor{mycolor2}\underline{39.3} 	 &\cellcolor{mycolor2}\underline{35.5} 	 &\cellcolor{mycolor2}\underline{36.7}  &\cellcolor{mycolor2}\underline{43.6}
\\
\multicolumn{1}{l|}{}&\cellcolor{mycolor3}$\bullet~$Ours$^\dagger$ ($\mathcal{BFS}=300$)   &\cellcolor{mycolor3}\textbf{53.0}  &\cellcolor{mycolor3}\textbf{51.1} 		 &\cellcolor{mycolor3}\textbf{52.2}	 &\cellcolor{mycolor3}\textbf{51.2}	 &\cellcolor{mycolor3}\textbf{50.2}	 &\cellcolor{mycolor3}\textbf{44.2} 	 &\cellcolor{mycolor3}\textbf{45.3} 	 &\cellcolor{mycolor3}\textbf{41.1} 	 &\cellcolor{mycolor3}\textbf{37.7} 	 &\cellcolor{mycolor3}\textbf{34.3}  &\cellcolor{mycolor3}\textbf{25.0} 	 &\cellcolor{mycolor3}\textbf{43.1}		 &\cellcolor{mycolor3}\textbf{37.5}	 &\cellcolor{mycolor3}\textbf{34.5} 	 &\cellcolor{mycolor3}\textbf{35.3}  &\cellcolor{mycolor3}\textbf{42.4}
\\        
                \bottomrule
            \end{tabular}
        }
    \end{threeparttable}
    \end{center}
\end{table*}

\section{More Ablation Studies}\label{sp-2}
All experiments in this section use the ResNet-50 with BatchNorm as the backbone.

\noindent\textbf{Hyper-Parameters}. We discuss the effect of several hyper-parameter variations and summarize them in Table~\ref{tab:ablation-hyper-parameter}: \ding{172} the trade-off parameter $\alpha$ in Eq.~5 of the main paper; \ding{173} the standard deviation level $\sigma$ in Gaussian noise perturbation.

As shown in Table~\ref{tab:ablation-hyper-parameter} \ding{172}, omitting this trade-off parameter $\alpha$ for adjusting the two dynamic weights results in suboptimal performance (\textit{e.g.}, $\alpha=0$), highlighting the effectiveness of the proposed gradient norm-based debiasing. Furthermore, a larger $\alpha$ leads to a lower average error rate, highlighting its effectiveness in refining these weights and enabling stable model adaptation during long-term distribution shifts.

As shown in Table~\ref{tab:ablation-hyper-parameter} \ding{173}, our method yields similar results when a small standard deviation is applied, indicating that slight perturbation is sufficient for selecting optimal samples. Conversely, a large standard deviation could significantly alter the model's predictions for samples that do not borders between the source- and target-domain data distributions, thereby weakening the discriminative power of our method.

\begin{table}[htp]
\caption{Performance comparisons on hyper-parameter variations, including \ding{172} the trade-off parameter $\alpha$, and \ding{173} the standard deviation level $\sigma$. The \textbf{best} and \underline{second-best} performances are highlighted.}
\label{tab:ablation-hyper-parameter}
\resizebox{1.0\linewidth}{!}{
\begin{tabular}{l|c|ccccc}
\toprule
     &Variants    &ImageNet-C  &ImageNet-R  &ImageNet-K  &ImageNet-A  &Avg. Err. \\  \toprule
    
 \multirow{5}{*}{\ding{172}}
 &\multicolumn{1}{c|}{0.0} &57.3 &52.6 &65.5 &99.0 &68.6 \\
 &\multicolumn{1}{c|}{0.2} &54.1  &51.7 &65.1  &\underline{98.2} &67.3 \\
 &\multicolumn{1}{c|}{0.4} &\underline{54.3}   &51.6  &\underline{64.5}  &98.4 &67.2 \\
 &\multicolumn{1}{c|}{0.6} &54.5  &\textbf{51.1}  &64.7  &\textbf{98.1} &\underline{67.1} \\
 &\multicolumn{1}{c|}{0.8} &\textbf{53.8}  &\underline{51.5}  &\textbf{64.3}  &\textbf{98.1} &\textbf{66.9} \\
\midrule

 \multirow{5}{*}{\ding{173}} 
 &\multicolumn{1}{c|}{0.01} &\underline{53.8}  &\underline{51.5}  &\textbf{64.3}  &\textbf{98.1} &\textbf{66.9} \\
 &\multicolumn{1}{c|}{0.02} &\textbf{53.7}  &\textbf{51.4}  &\textbf{64.3}  &\underline{98.3} &\textbf{66.9} \\
 &\multicolumn{1}{c|}{0.03} &54.1  &51.8  &64.9  &98.5 &\textbf{66.9} \\
 &\multicolumn{1}{c|}{0.1} &\underline{53.8}  &51.6  &\underline{64.7}  &\textbf{98.1} &\underline{67.1} \\
 &\multicolumn{1}{c|}{1.0} &54.6  &52.9  &66.8  &98.0 &68.1 \\
\bottomrule
\end{tabular}}
\end{table}

\section{Dataset Details}\label{sp-3}

\textbf{ImageNet-C}. ImageNet-C~\cite{hendrycks2019benchmarking} is a dataset derived from the validation set of the original ImageNet with common corruptions and perturbations, such as `Gaussian Noise', `Shot Noise', `Impulse Noise', `Defocus Blur', `Glass Blur', `Motion Blur', `Zoom Blur', `Snow', `Frost', `Fog', `Brightness', `Contrast', `Elastic Transform', `Pixelate', and `JPEG Compression'. Each corruption type is applied at five levels of severity, resulting in 50,000 images per corruption type. Overall, the dataset comprises 750,000 images across 1,000 classes.

\noindent \textbf{ImageNet-Rendition (R)}. ImageNet-R~\cite{hendrycks2021many} is a dataset with diverse artistic renditions, such as cartoons, paintings, origami, embroidery, toys, sculptures, and so on. It features renditions of 200 ImageNet classes, comprising a total of 30,000 images. 

\noindent \textbf{ImageNet-Sketch (K)}. ImageNet-K~\cite{wang2019learning} is a dataset designed to provide sketch-based representations of objects belonging to the ImageNet database. It consists of hand-drawn sketches corresponding to 50,000 images from 1,000 different categories in ImageNet.

\noindent \textbf{ImageNet-A}. ImageNet-A~\cite{hendrycks2021nae} provides natural adversarial examples that are challenging for models to recognize correctly while still being visually similar to the original classes. It contains 7,500 images across 200 categories. Each category corresponds to a class from the original ImageNet dataset.

\noindent \textbf{PACS}. PACS~\cite{li2017deeper} includes a total of 9,991 images across four domains, such as `Photo', `Art Painting', `Cartoon' and `Sketch'. Each domain contains seven categories.

\section{More Experimental Details}\label{sp-4}

\subsection{Pre-training Protocol on PACS} We employ the ResNet-18 with pre-trained weights, specifically `ResNet18\_Weights.DEFAULT' from PyTorch. Following~\cite{gui2024active} and~\cite{gulrajani2020search}, we fix the statistics in the batch normalization layers in the pre-trained model. We set the batch size to 32 and train the model for 40 epochs using the Adam optimizer, with a learning rate of 0.0001 and a weight decay of 5e-5.

\subsection{Implementation Details of TTA and ATTA Baselines} 
\noindent\textbf{TENT}. For TENT~\cite{wang2020tent}, we use the SGD optimizer with a learning rate of 0.00025 and a momentum of 0.9 on ImageNet-C, -R, -K, and -A. We use the Adam optimizer with a learning rate of 0.005 on PACS. The implementation follows the official code\footnote{\href{1}{https://github.com/DequanWang/tent}}.

\noindent\textbf{CoTTA}. For CoTTA~\cite{wang2022continual}, we use the SGD optimizer with a learning rate of 0.01 and a momentum of 0.9 on ImageNet-C, -R, -K, and -A. And the restoration factor, exponential moving average factor, the average probability threshold, and the augmentation number are set to 0.01, 0.999, 0.1, 32, respectively. Moreover, we use the Adam optimizer with a learning rate of 0.01 on PACS. And the restoration factor, exponential moving average factor, the average probability threshold, and the augmentation number are set to 0.01, 0.999, 0.72, 32, respectively.
The implementation follows the official code\footnote{\href{1}{https://github.com/qinenergy/cotta}}.

\noindent\textbf{ETA}. For ETA~\cite{niu2022efficient}, we use the SGD optimizer with a learning rate of 0.00025 and a momentum of 0.9 on ImageNet-C, -R, -K, and -A. Moreover, we use the Adam optimizer with a learning rate of 0.001 on PACS. We set the exponential moving average factor, the cosine similarity threshold, and the entropy threshold to 0.9, 0.05, and $0.4 \times \ln(\mathcal{C})$, respectively. Here, $\mathcal{C}$ is the number of classes.
The implementation follows the official code\footnote{\href{1}{https://github.com/mr-eggplant/EATA}}.

\noindent\textbf{SAR}. For SAR~\cite{niu2023towards}, we use the SAM optimizer with a learning rate of 0.001 and a momentum of 0.9 on ImageNet-C, -R, -K, and -A. Moreover, 
we use the Adam optimizer with a learning rate of 0.001 on PACS. We set the reset factor, the entropy threshold, and the exponential moving average factor to 0.2, $0.4 \times \ln(\mathcal{C})$, and 0.9, respectively, for all datasets. The implementation follows the official code\footnote{\href{1}{https://github.com/mr-eggplant/SAR}}.

\noindent\textbf{SimATTA}. For SimATTA~\cite{gui2024active}, we employ the SGD optimizer with a learning rate of 0.00025 and a momentum of 0.9 on ImageNet-C, -R, -K, and -A databases. For PACS, we use the Adam optimizer with a learning rate of 0.005. The maximum length of anchors is set to 50, and the entropy threshold is set to $0.4 \times \ln(\mathcal{C})$. This adjustment to the entropy threshold is necessary because the original threshold is not appropriate for ImageNet-C, leading to suboptimal performance. The buffer size is fixed to 300 for fair comparison. The implementation follows the official code\footnote{\href{1}{https://github.com/divelab/ATTA}}.

\noindent\textbf{CEMA}. For CEMA~\cite{chen2024towards}, we employ the SGD optimizer with a learning rate of 0.00025 and a momentum of 0.9 on ImageNet-C, -R, -K, and -A databases.
And the maximum entropy threshold, the minimum entropy threshold, and the decreasing factor are set to $0.4 \times \ln(\mathcal{C})$, $0.02 \times \ln(\mathcal{C})$, and 1.0, respectively. The buffer size is set to 300 for fair comparison. The implementation follows the official code\footnote{\href{1}{https://github.com/chenyaofo/CEMA}}.

\noindent\textbf{HILTTA}. For HILTTA~\cite{li2024exploring}, we use the experimental results reported in the original paper.

\noindent\textbf{Baseline}. The Baseline method, which builds on TENT, randomly selects a specified number of samples from each online batch for manual annotation and then performs ATTA using Eq. 2 of the main paper. We employ the SGD optimizer with a learning rate of 0.00025 and a momentum of 0.9 on ImageNet-C, -R, -K, and -A databases.  We use the Adam optimizer with a learning rate of 0.005 on PACS.
\end{document}